\documentclass[10pt,twocolumn,letterpaper]{article}

\usepackage[pagenumbers]{cvpr} 

\definecolor{cvprblue}{rgb}{0.21,0.49,0.74}
\usepackage[pagebackref,breaklinks,colorlinks,allcolors=cvprblue]{hyperref}

\usepackage{graphicx}
\usepackage{booktabs}
\usepackage{multirow}
\usepackage{makecell}
\usepackage{multirow}
\usepackage{adjustbox}
\usepackage{colortbl}
\usepackage{amssymb}  
\usepackage{pifont}
\usepackage{afterpage}
\usepackage{mathrsfs}
\usepackage[hang,flushmargin]{footmisc}
\usepackage{textcomp}
\usepackage{tcolorbox}
\usepackage{bbding, fontawesome}
\usepackage{graphicx}
\usepackage{stfloats} 
\usepackage{pifont}

\def\paperID{2097} 
\def\confName{CVPR}
\def\confYear{2026}

\title{SoccerMaster: A Vision Foundation Model for Soccer Understanding}

\author{
    Haolin Yang$^{1,2}$,\, Jiayuan Rao$^{1}$,\, Haoning Wu$^{1}$,\, Weidi Xie$^{1}$ \\[4pt]
    $^{1}$School of Artificial Intelligence, Shanghai Jiao Tong University \\
    $^{2}$Shanghai Innovation Institute\\[4pt]
    \url{https://haolinyang-hlyang.github.io/SoccerMaster}
}

\begin{document}

\twocolumn[{%
\renewcommand\twocolumn[1][]{#1}%
\maketitle
\vspace{-21pt}
\begin{center}
   \centering
   \includegraphics[width=\textwidth]{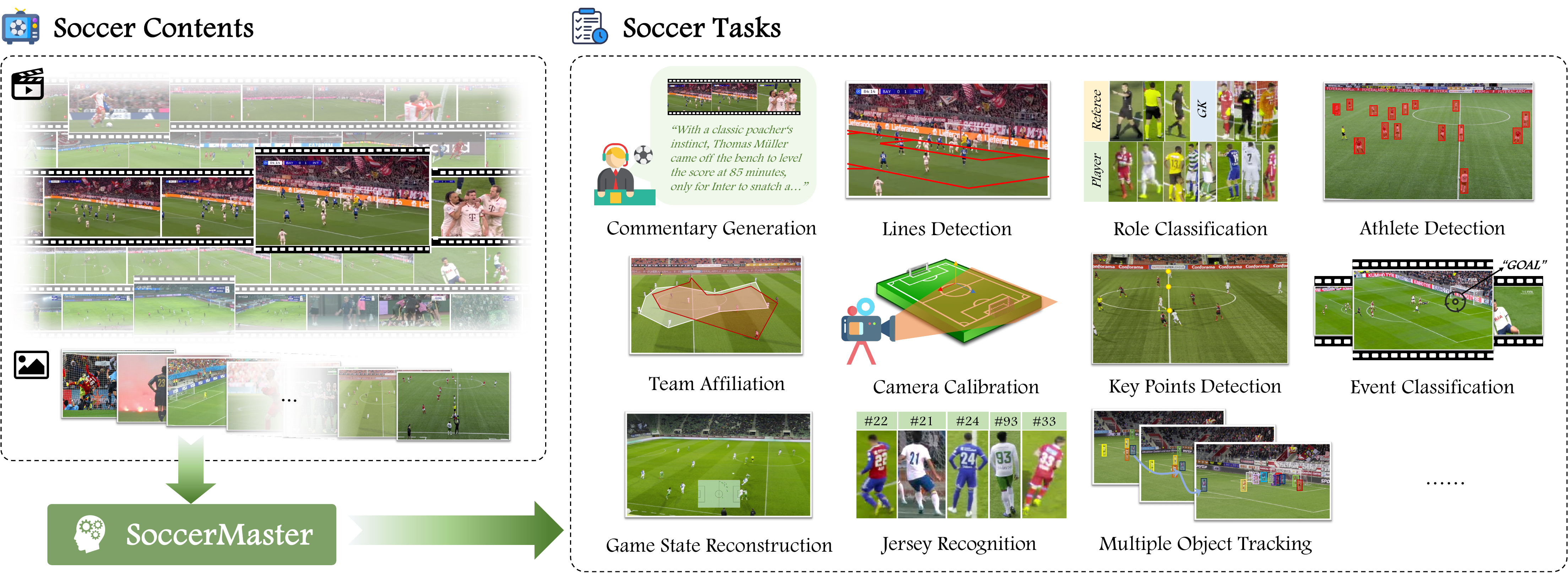}
   \vspace{-12pt}
   \captionsetup{hypcap=false}
   \captionof{figure}{
        \textbf{SoccerMaster} is a unified soccer-specific vision foundation model that leverages diverse soccer content, including images and videos, to support a wide range of soccer understanding tasks, such as commentary generation, detection, tracking, classification, {\em etc}.
   }
  \label{fig:teaser}
 \end{center}
}]

\begin{abstract}

Soccer understanding has recently garnered growing research interest due to its domain-specific complexity and unique challenges.
Unlike prior works that typically rely on isolated, task-specific expert models, this work aims to propose a unified model to handle diverse soccer visual understanding tasks, ranging from fine-grained perception~({\em e.g.}, athlete detection and identification) to high-level semantic reasoning~({\em e.g.}, event classification). 
Concretely, our contributions are threefold: 
(i) we present \textbf{SoccerMaster}, the first soccer-specific vision foundation model that unifies diverse tasks within a single framework via \textbf{supervised multi-task pretraining};
(ii) we develop an automated data curation pipeline, \textbf{SoccerFactory}, to generate scalable spatial annotations, and integrate multiple existing soccer video datasets as a comprehensive pretraining data resource for multi-task pretraining;
and 
(iii) we conduct extensive evaluations demonstrating that \textbf{SoccerMaster} consistently outperforms task-specific expert models across diverse downstream tasks, highlighting its breadth and superiority. 
The data, code, and model will be publicly available.

\end{abstract}
    
\section{Introduction}
\label{sec:introduction}
Soccer, as one of the world's most popular sports, enjoys a vast global audience and generates substantial commercial and cultural impact. 
Such popularity has driven growing research interest in soccer visual understanding~\cite{GOAL, rao2025unisoccer, rao2025socceragent, SoccerNetv3, SoccerNetv2, SoccerNet}. 
However, the sport's inherent dynamism, complex rules, and varying broadcast conditions present unique challenges for computer vision systems. 

A central dichotomy in soccer understanding lies between \textit{spatial perception} and \textit{semantic reasoning}. 
The former requires fine-grained geometric precision for tasks such as athlete detection~\cite{SoccerNet-GSR, Vandeghen2022SemiSupervised}, tracking~\cite{SoccerNetv3-Tracking, Magera2025BroadTrack, prtreid}, jersey recognition~\cite{JerseyNumber, balaji2023jersey}, and camera calibration~\cite{CameraCalibration, SoccerNet-Camera}. 
The latter demands high-level abstraction for action spotting~\cite{SoccerNet}, event classification~\cite{SoccerNetv2}, foul recognition~\cite{held2023vars, held2024XVARS}, and commentary generation~\cite{densecap, rao2024matchtimeautomaticsoccergame, GOAL}. 
Current paradigms predominantly rely on fragmented, task-specific expert models~\cite{rao2024matchtimeautomaticsoccergame, held2024XVARS, SoccerNetv3-Tracking} to address the challenges in isolation. 
While recent initiatives~\cite{rao2025unisoccer} have attempted to unify these tasks via vision-language alignment, they often neglect dense spatial objectives during pretraining. 
This omission results in models that can describe \textit{what} is happening but struggle to pinpoint \textit{where} or \textit{who} is involved, leading to a disconnect between spatial perception and semantic reasoning.

To address the aforementioned challenges, we introduce \textbf{SoccerMaster}, the first soccer-specific vision foundation model that unifies fine-grained spatial perception and high-level semantic representations within a single framework.
As depicted in Fig.~\ref{fig:teaser}, \textbf{SoccerMaster} is designed to learn holistic representations through multi-task learning. 
By simultaneously optimizing for (i) athlete detection and identification, (ii) pitch registration, (iii) event classification, and (iv) vision–language alignment, we enforce a shared representation that captures both the geometry of the pitch and the semantics of the game.

Large-scale training requires massive data. 
To address the scarcity of dense spatial labels, we develop \textbf{SoccerFactory}, an automated data curation pipeline capable of generating high-quality annotations from broadcast footage at scale. 
By integrating the data curated by \textbf{SoccerFactory} with existing datasets, we construct a comprehensive pretraining resource. 
This resource effectively balances \textit{breadth}~(coverage of diverse events, leagues, and camera styles) with \textit{depth}~(dense spatial labels and temporally precise semantics), driving effective multi-task optimization.

We conduct extensive evaluations across a broad spectrum of tasks, spanning both spatial perception~({\em e.g.}, athlete detection, pitch registration, camera calibration, multiple object tracking) and semantic reasoning~({\em e.g.}, vision-language alignment, event classification, commentary generation).
Our results demonstrate that \textbf{SoccerMaster} substantially outperforms both general-purpose vision foundation models~({\em e.g.}, SigLIP~2~\cite{tschannen2025siglipv2}, DINOv3~\cite{simeoni2025dinov3}) and soccer-specific models~(MatchVision~\cite{rao2025unisoccer}) across all pretraining tasks with direct evaluation. 
Moreover, with only lightweight task-specific fine-tuning, our model establishes state-of-the-art or competitive results on downstream applications such as camera calibration~\cite{SoccerNet2022, CameraCalibration}, multiple object tracking~\cite{SoccerNetv3-Tracking}, and commentary generation~\cite{rao2024matchtimeautomaticsoccergame}.
These findings validate that \textbf{SoccerMaster} has learned universal, transferable representations that effectively generalize to diverse soccer visual understanding scenarios.

Overall, our contributions can be summarized as follows:
(i) we introduce \textbf{SoccerMaster}, the first soccer-specific vision foundation model that unifies diverse soccer visual understanding tasks within a single framework, serving as a versatile backbone for comprehensive soccer analysis;
(ii) we introduce a supervised multi-task pretraining strategy that jointly optimizes fine-grained spatial perception and high-level semantic reasoning tasks on soccer videos, enabling the model to learn multi-granularity representations that capture both local spatial details and global semantic context;
(iii) we develop \textbf{SoccerFactory}, an automated data curation pipeline to generate large-scale spatial annotations from broadcast footage, and integrate these with various existing soccer video datasets to construct a comprehensive soccer-specific pretraining resource;
and
(iv) we conduct extensive evaluations demonstrating that \textbf{SoccerMaster} can serve as a foundation model for soccer understanding, which achieves state-of-the-art or competitive performance on all downstream tasks through simple fine-tuning.
\section{Related Work}
\label{sec:related_work}
\noindent \textbf{Vision foundation models}~(VFMs)~\cite{ViT, zhai2022scaling, clip, tschannen2025siglipv2, dinov2, simeoni2025dinov3, SAM2, zheng2025one, liu2026versavit, yan2026omnistream, yan2025learning} have revolutionized computer vision by learning universal visual representations through large-scale pretraining, demonstrating strong generalization and transfer capabilities across diverse tasks.
This has further inspired their adaptation and application in numerous downstream vertical domains or tasks, such as image captioning~\cite{blip2, li2022blip}, 3D perception~\cite{wang2025vggt, DepthAnythingv2}, medical imaging~\cite{zhou2024knowledge, wu2023generalist, wu2025mrgen}, robotics~\cite{firoozi2025foundation}, and autonomous driving~\cite{gao2024survey}. 
However, their exploration in video domains, especially in highly dynamic and professional sports visual understanding, remains relatively nascent.

\vspace{2pt}
\noindent \textbf{Sports analysis}~\cite{thomas2017computer} has gradually become a rapidly evolving field with commercial and research value.
While the general sports artificial intelligence~\cite{wu2022sportsvideoanalysislargescale, li2024sports, xia2024sportqa, xia2024sportu} remains under exploration, recent research on specific sports understanding tasks has achieved significant progress, including action spotting~\cite{li2021multisports}, automated scoring~\cite{xu2022finediving, shao2020finegym, BASKET_CVPR25, sportsassessment}, commentary generation~\cite{xi2025simple, xi2025eika, Xi_2025_ICCV, wu2022sportsvideoanalysislargescale}, activity recognition~\cite{MLB, Ma2021NPURD,7780586}, tracking~\cite{8733019, cui2023sportsmot}, and game state analysis~\cite{liu2025f, mao2023leapfrog}. 
Among these research efforts, soccer, as a globally popular sub-domain with rich spatiotemporal dynamics, has received widespread attention.

\vspace{2pt}
\noindent \textbf{Soccer Datasets.} 
The rapid advancement of visual soccer understanding owes much to the continuous development of large-scale, high-quality datasets. 
Notably, the \textbf{SoccerNet} series~\cite{SoccerNet, SoccerNetv2, SoccerNet2022, SoccerNet2023, cioppa2024soccernet2024challengesresults, giancola2025soccernet2025challengesresults} stands as the indispensable cornerstone and the most widely used benchmark in this field. 
Evolving from its initial focus on action spotting~\cite{SoccerNet}, SoccerNet has expanded into a comprehensive suite that supports a myriad of fundamental tasks, including camera shot segmentation and boundary detection~\cite{SoccerNetv2}, dense video captioning~\cite{densecap}, multi-view foul recognition~\cite{held2023vars}, camera calibration~\cite{CameraCalibration}, tracking~\cite{SoccerNetv3-Tracking}, re-identification~\cite{SoccerNet2022}, jersey number recognition~\cite{SoccerNet2023}, game state reconstruction~\cite{SoccerNet-GSR}, and monocular depth estimation~\cite{leduc2024soccernet}. 
Beyond SoccerNet, large-scale datasets like SoccerReplay-1988~\cite{rao2025unisoccer} further enrich the domain with rich annotations like event labels and textual commentaries. 
While these pioneering datasets establish a rich foundation for the community, their annotations are predominantly tailored for distinct, decoupled tasks. 
Furthermore, there remains a noticeable scarcity of large-scale datasets that provide dense, spatial annotations specifically from the main-camera perspective. 
To train a robust and unified vision foundation model, there remains a critical need for an automated and scalable data curation pipeline capable of generating unified, fine-grained spatial annotations. 
This motivates our introduction of the \textbf{SoccerFactory} pipeline to construct our pretraining dataset. 

\vspace{2pt}
\noindent \textbf{Soccer visual understanding} involves diverse tasks~\cite{cioppa2024soccernet2024challengesresults, giancola2025soccernet2025challengesresults}, while early research predominantly focuses on building specific models to individually address particular tasks, such as athlete detection~\cite{Vandeghen2022SemiSupervised, SoccerNet-GSR}, tracking~\cite{cui2023sportsmot, prtreid}, jersey number recognition~\cite{SoccerNet-GSR, balaji2023jersey}, camera calibration~\cite{gutierrez2024pnlcalib, falaleev2024enhancing}, event classification~\cite{SoccerNet, SoccerNetv2}, commentary generation~\cite{densecap, rao2024matchtimeautomaticsoccergame, GOAL, youling} and foul recognition~\cite{held2023vars, held2024XVARS}.
This fragmented paradigm inevitably increases the exploration cost in soccer understanding. 
Although recent work~\cite{rao2025socceragent} has attempted to leverage multimodal large language models~(MLLMs) to handle multiple tasks simultaneously, they only focus on semantic-level tasks such as question-answering. 
These limitations inspire us to build \textbf{SoccerMaster}, a vision foundation model for the soccer domain that can unify fine-grained spatial perception and high-level semantic reasoning tasks in soccer within a single framework.
\section{Task Specifications}
\label{subsec:task_specifications}
To develop and evaluate our vision foundation model, we define a set of tasks over a given video clip, $\mathcal{V} = \{\mathcal{I}_1, \dots \mathcal{I}_T\}$, 
where $\mathcal{I}_t \in \mathbb{R}^{H \times W \times 3}$ denotes a RGB frame. 
These tasks involve both fine-grained {\em spatial perception} and high-level {\em semantic reasoning}, and are categorized into two groups: 
\textbf{pretraining tasks} that are used to jointly train \textbf{SoccerMaster}, including
(i) athlete detection and identification, 
(ii) pitch registration, 
(iii) event classification, 
(iv) vision-language alignment,
and \textbf{downstream tasks} that demonstrate the model's versatility and generalization ability, including
(v) commentary generation,
(vi) camera calibration,
(vii) multiple object tracking.
Details of these tasks are presented as follows.

\vspace{2pt} 
\noindent \textbf{Athlete detection and identification.}
In soccer game analysis, a fundamental task lies in accurately localizing and identifying the athletes.
For each frame~($\mathcal{I}_t$), we aim to detect a set of athlete instances~($\mathcal{A}_t = \{ \mathbf{a}_i \mid i=1,\dots, N_a \} \in \mathbb{R}^{N_a \times 6}$), where each instance is characterized by $\mathbf{a}_i = (\mathbf{b}_i; \mathbf{r}_i; \mathbf{n}_i)$.
Here, $\mathbf{b}_i = (x_i, y_i, w_i, h_i) \in \mathbb{R}^4$ represents the 2D bounding box of the detected athlete; $\mathbf{r}_i \in \{\mathrm{goalkeeper}, \mathrm{player}, \mathrm{referee}\}$ denotes the athlete's role, 
and $\mathbf{n}_i \in \mathbb{N}$ indicates the jersey number. Notably, when the detected athlete's role~($\mathbf{r}_i$) is $\mathrm{goalkeeper}$ or $\mathrm{referee}$, or when the jersey number is not visible due to occlusion or viewpoint issues, the corresponding number~($\mathbf{n}_i$) takes the value $\texttt{null}$.

\vspace{2pt}
\noindent \textbf{Pitch registration.}
Another key spatial perception task lies in detecting field keypoints and lines, which serves as the cornerstone for camera calibration~\cite{gutierrez2024pnlcalib, falaleev2024enhancing}.
The objective is to identify and localize structural elements of the soccer pitch, including keypoints~({\em e.g.}, intersection between side lines) and lines~({\em e.g.}, middle line and goal crossbar).
For each frame~($\mathcal{I}_t$), this task expects to output two sets:
\begin{align}
    \mathcal{K}_t & = \{ \mathbf{k}_i = (x_i, y_i; t_i) \mid i=1,\dots,N_k \} \in \mathbb{R}^{N_k \times 3} \\
    \mathcal{L}_t & = \{ \mathbf{l}_j = (\mathbf{p}_{j}^{a}, \mathbf{p}_{j}^{b}; t_j) \mid j=1,\dots,N_l \} \in \mathbb{R}^{N_l \times 5}
\end{align}
where $\mathbf{k}_i$ represents a keypoint with coordinates~$(x_i, y_i)$ and type ($t_i \in \mathcal{T}_{\mathrm{keypoint}}$), while $\mathbf{l}_j$ denotes a line segment determined by two unordered endpoints~$\mathbf{p}_{j}^{a} = (x_{j}^{a}, y_{j}^{a})$ and $\mathbf{p}_{j}^{b} = (x_{j}^{b}, y_{j}^{b})$ with line type~($t_j \in \mathcal{T}_{\mathrm{line}}$). 
Here, candidate sets $\mathcal{T}_{\mathrm{keypoint}}$ and $\mathcal{T}_{\mathrm{line}}$ follow the definition in~\cite{gutierrez2024pnlcalib}.

\vspace{2pt} 
\noindent \textbf{Event classification.}
As a core semantic reasoning task in soccer understanding, 
event classification aims to recognize key actions within video segments. 
Given a video clip~($\mathcal{V}$), the model predicts an event label, 
$\mathbf{e} \in \mathcal{E}$, where $\mathcal{E}$ denotes the set of 24 event categories defined in SoccerReplay-1988 dataset~\cite{rao2025unisoccer}, including {\em goal}, {\em corner}, {\em yellow card}, etc. 

\vspace{2pt} 
\noindent \textbf{Vision-language alignment.}
This semantic reasoning task aims to learn semantic correspondences between video content and textual commentary. Concretely, the objective is to align the video semantic features with the text features extracted by a pretrained text encoder~({\em e.g.} SigLIP~2~\cite{tschannen2025siglipv2}), 
where their embeddings possess a high similarity score in a shared embedding space.

\vspace{2pt} 
\noindent \textbf{Commentary generation.}
Building upon vision-language alignment pretraining, commentary generation aims to generate a textual commentary~($\hat{\mathcal{T}}$) of the events and actions occurring in a soccer video clip~($\mathcal{V}$), following~\cite{rao2025unisoccer, rao2024matchtimeautomaticsoccergame}.

\vspace{2pt} 
\noindent \textbf{Camera calibration.}
This task aims to estimate the geometric relationship between the 2D image plane and the 3D soccer field. 
Formally, given an input frame ($\mathcal{I} \in \mathbb{R}^{H \times W \times 3}$), the task is to predict the camera intrinsics~($\mathbf{K} \in \mathbb{R}^{3 \times 3}$) and extrinsics~($[\mathbf{R}|\mathbf{t}]$), where $\mathbf{R} \in \mathbb{R}^{3 \times 3}$ is the rotation matrix and $\mathbf{t} \in \mathbb{R}^{3}$ is the translation vector.
This task leverages the field keypoints and lines detected through pitch registration as intermediate representations.

\vspace{2pt} 
\noindent \textbf{Multiple object tracking.}
This task extends athlete detection to establish temporal identity consistency across video frames. 
Beyond predicting bounding boxes for each athlete in each frame, it assigns a unique identity~($\mathrm{id}_i \in \mathbb{N}$) to each detection, thereby constructing trajectories that maintain consistent identities throughout the video.

Training such a foundation model for diverse soccer understanding tasks demands data spanning both spatial perception and semantic reasoning. 
While existing datasets cover semantic tasks~\cite{SoccerNetv2, rao2024matchtimeautomaticsoccergame, rao2025unisoccer}, fine-grained spatial annotations~\cite{SoccerNet-GSR} remain limited and costly to scale up.
\begin{figure*}[t]
    \centering
    \includegraphics[width=\linewidth]{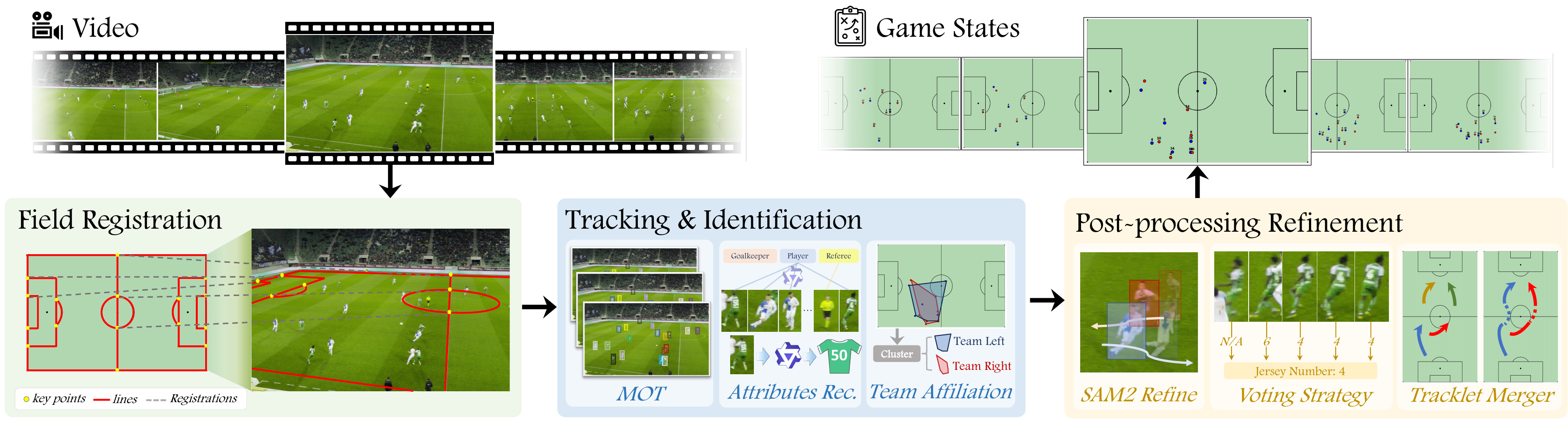}
    \vspace{-6pt}
    \caption{
        \textbf{Automated Data Curation Pipeline.}
        Our pipeline processes input videos through three stages: 
        (i) field registration establishes geometric correspondences between image and canonical pitch coordinates via keypoint detection; 
        (ii) tracking and identification transforms frames into athlete trajectories through detection, role and team classification, and ReID-based tracking; 
        and 
        (iii) post-processing refinement improves tracking accuracy through SAM2-based segmentation and ensures temporal consistency via majority voting. 
    }
    \label{fig:gsr_pipeline}
    \vspace{-6pt}
\end{figure*}

\section{Pretraining Dataset}
\label{sec:dataset}
To address the scarcity of large-scale spatial annotations, we have developed \textbf{SoccerFactory}, an automated data curation pipeline that extracts labels directly from broadcast footage. 
By combining these automatically generated annotations with existing resources, we construct a comprehensive dataset that supports unified pretraining for both spatial perception and semantic reasoning tasks. 
This section introduces the automated annotation pipeline in Sec.~\ref{subsec:automated_data_curation}, evaluates its quality in Sec.~\ref{subsec:pipeline_performance}, and presents the composition of the integrated dataset in Sec.~\ref{subsec:data_integration_and_statistics}.

\subsection{The SoccerFactory Pipeline}
\label{subsec:automated_data_curation}
We introduce \textbf{SoccerFactory}, an automated data curation pipeline designed to transform raw broadcast footage into structured annotations for spatial perception tasks.
The pipeline consists of three core stages: field registration, tracking, and identification, as well as post-processing refinement, as illustrated in Fig.~\ref {fig:gsr_pipeline}.

\vspace{2pt} 
\noindent \textbf{Field registration.} 
Following~\cite{SoccerNet-GSR}, we establish geometric correspondences between image and pitch coordinates through field keypoint and line detectors.
The detected keypoints and lines are then processed using the PnL module~\cite{gutierrez2024pnlcalib} to estimate camera parameters, enabling projection between image and standardized pitch coordinates. This further allows refining the detected keypoints and lines as training annotations via projection from the canonical pitch.

\vspace{2pt} 
\noindent \textbf{Tracking and identification.} 
We employ YOLOv8~\cite{varghese2024yolov8}, specifically fine-tuned on soccer data, to detect 
{\em players}, {\em goalkeepers}, and {\em referees}, followed by StrongSORT~\cite{du2023strongsort} tracking with ReID embeddings extracted by PRTReID~\cite{prtreid}. 
For each detection, we crop the bounding box region and apply Qwen2.5-VL~\cite{Qwen2.5-VL} to recognize roles and jersey numbers, 
which are further filtered by a legibility classifier~\cite{JerseyNumber}.
Team affiliation is determined through clustering on tracklet-averaged ReID embeddings and pitch coordinates mapped via the estimated camera parameters.

\vspace{2pt} 
\noindent \textbf{Post-processing refinement.} 
We enhance annotation quality through SAM2-based~\cite{SAM2} segmentation refinement to recover missed detections from the detector and correct tracking identity switches. 
Additionally, we use majority voting across tracklets for jersey numbers and role classifications to improve temporal consistency. 
To further consolidate fragmented trajectories, we merge short tracklet fragments into longer tracklets by leveraging both ReID embeddings and jersey number consistency, similar to~\cite{golovkin2025broadcast}.

The pipeline generates comprehensive annotations including: athlete bounding boxes with roles, team affiliations, and jersey numbers; soccer field keypoints and lines; camera parameters; and player trajectories in pitch coordinates. 
These automatically generated annotations enable scalable data production for spatial perception tasks.

\subsection{Quality Assessment of SoccerFactory}
\label{subsec:pipeline_performance}
To assess the effectiveness of \textbf{SoccerFactory}, we apply it to the official test set of the SoccerNet Game State Reconstruction~(GSR) Challenge~\cite{giancola2025soccernet2025challengesresults}, which provides a comprehensive assessment across multiple dimensions, covering athlete tracking, role classification, jersey number recognition, camera calibration, and team affiliation.
As shown in Tab.~\ref{tab:performance_gsr_pipeline}, results from our pipeline outperform the leading methods, with a superior GS-HOTA score of \textbf{64.1}, demonstrating its capability to generate high-quality annotations at scale, offering a practical alternative to expensive manual labeling.

\begin{table}[!t]
    \caption{
        \textbf{Comparison on GSR.} 
        SoccerMaster outperforms the challenge winner KIST-GSR on the SoccerNet-GSR test set. The best and second-best results are \textbf{bolded} and \underline{underlined}.}
    \vspace{-3pt}
    \centering
    \footnotesize
    \resizebox{.86\linewidth}{!}{
        \begin{tabular}{l | c c c}
        \toprule
        \textbf{Method} & \textbf{GS-HOTA} & \textbf{GS-DetA} & \textbf{GS-AssA} \\
        \midrule
        Playbox \& MIXI & 58.1 & 41.3 & \textbf{81.6} \\
        Metrica-Sports & 58.2 & 44.4 & 76.2 \\
        KIST-GSR & \underline{61.5} & \underline{48.5} & 78.0 \\
        \midrule
        \textbf{Ours} & \textbf{64.1} & \textbf{51.5} & \underline{79.9} \\
        \bottomrule
        \end{tabular}
    }
    \vspace{-9pt}
    \label{tab:performance_gsr_pipeline}
\end{table}

\subsection{Pretraining Dataset Composition}
\label{subsec:data_integration_and_statistics}

As detailed in Tab.~\ref{tab:dataset_composition}, we integrate automatically generated annotations from \textbf{SoccerFactory} with multiple existing soccer datasets to construct a comprehensive, large-scale pretraining dataset. 
For spatial perception tasks, our pipeline processes main-camera clips from 500 matches in SoccerNet-v2~\cite{SoccerNetv2}. 
To ensure stable and comprehensive field coverage, we exclude broadcast views with incomplete visibility, such as replays, close-ups, and alternative angles. 
These automatically generated annotations are combined with manually annotated data from SoccerNet-GSR~\cite{SoccerNet-GSR}, where tracklet-level jersey number annotations are converted to frame-level labels using a legibility filter~\cite{JerseyNumber} to mitigate occlusion ambiguity.
For semantic reasoning tasks, we incorporate large-scale videos~(sampled at 1FPS) from SoccerNet-v2~\cite{SoccerNetv2}, MatchTime~\cite{rao2024matchtimeautomaticsoccergame}, and SoccerReplay-1988~\cite{rao2025unisoccer}.
Overall, our pretraining dataset comprises about 7.45M frames across 248.3K video segments, with 2.75M frames for spatial perception and 4.71M frames for semantic reasoning, providing extensive coverage for unified pretraining across diverse soccer understanding tasks.
Further details regarding the data are provided in Sec.~\ref{sec:soccerfactory_dataset_construction} of the \textbf{Appendix}.
\section{Models}
\label{sec:archi}
\begin{figure*}[t]
    \centering
    \includegraphics[width=\linewidth]{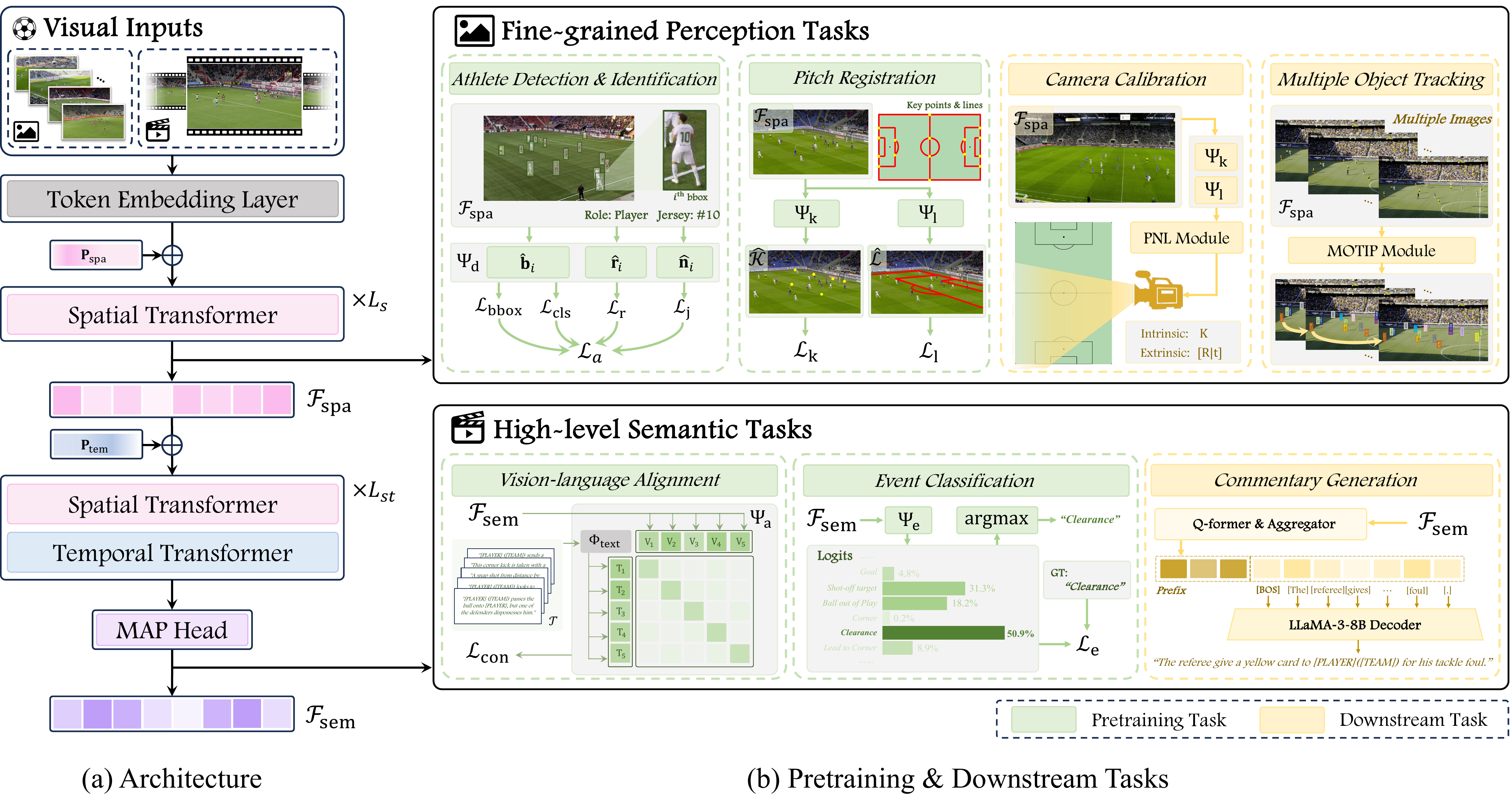}
    \vspace{-12pt}
    \caption{
      \textbf{SoccerMaster Architecture.} 
      (a) The architecture of SoccerMaster, which encodes both soccer videos and images through spatial and temporal attention modules to generate semantically rich representations. 
      (b) The pretraining tasks and downstream adaptations of SoccerMaster across both spatial perception and semantic understanding tasks.
   }
   \vspace{-9pt}
\label{fig:architecture}
\end{figure*}

Building on the task definitions in Sec.~\ref{subsec:task_specifications} and the dataset overview in Sec.~\ref{sec:dataset}, this section elaborates on the architecture, training, and evaluation of SoccerMaster, with an overview illustrated in Fig.~\ref{fig:architecture}.
We begin with problem formulations in Sec.~\ref{subsec:problem_formulation}, followed by the model architecture in Sec.~\ref{subsec:architecture} and multi-task pretraining strategy in Sec.~\ref{subsec:multitask_pretraining}. Finally, Sec.~\ref{subsec:downstream_tasks} describes how the model is adapted to various downstream tasks.

\subsection{Problem Formulation}
\label{subsec:problem_formulation}
Our soccer vision foundation model, \textbf{SoccerMaster}, consists of a visual encoder~($\Phi_{\mathrm{enc}}$) and multiple task-specific heads that enable diverse soccer visual understanding capabilities.
Given a soccer video segment~($\mathcal{V} = \{\mathcal{I}_1, \dots \mathcal{I}_T\} \in \mathbb{R}^{T \times 3 \times H \times W}$), our visual encoder~($\Phi_{\mathrm{enc}}$) first extracts spatial features~($\mathcal{F}_{\mathrm{spa}}$) and semantic features~($\mathcal{F}_{\mathrm{sem}}$) embedded in the video, which can be formulated as:
\begin{align}
    \mathcal{F} = \{\mathcal{F}_{\mathrm{spa}}, \mathcal{F}_{\mathrm{sem}}\} = \Phi_{\mathrm{enc}}(\mathcal{V})
\end{align}
Leveraging the extracted features~($\mathcal{F}$), our model tackles the aforementioned tasks via corresponding task-specific heads~($\Psi_{\mathrm{out}}$), with the complete set of outputs unified as:
\begin{align}
    \{\mathcal{A}, \{\mathrm{id}_i\}, \mathcal{K}, \mathcal{L}, \mathbf{K}, \mathbf{R}, \mathbf{t}, \mathbf{e}, \hat{\mathcal{T}}\} = \Psi_{\mathrm{out}}(\mathcal{F})
\end{align}
For vision-language alignment, the alignment head~($\Psi_{\mathrm{a}}$) computes the similarity~($\mathbf{s}$) between video semantic features~($\mathcal{F}_{\mathrm{sem}}$) and text features of textual commentary~($\mathcal{T}$), extracted by a pretrained text encoder~($\Phi_{\mathrm{text}}$), denoted as:
\begin{align}
    \mathbf{s} = \Psi_{\mathrm{a}}(\mathcal{F}_{\mathrm{sem}}, \Phi_{\mathrm{text}}(\mathcal{T}))
\end{align}

\begin{table}[t]
    \caption{
        \textbf{Pretraining Dataset Composition.} 
        Our pretraining dataset comprises approximately 7.45M frames across 248.3K video segments, with 2.75M frames for spatial perception tasks and 4.71M frames for semantic reasoning tasks (sampled at 1FPS).
    }
    \vspace{-3pt}
\centering
\footnotesize
    \resizebox{0.98\columnwidth}{!}{
        \begin{tabular}{l|c|l}
        \toprule
        \textbf{Dataset} & \textbf{\# Frames} & \textbf{Pretraining Tasks} \\
        \midrule
        \rowcolor{gray!20}
        \multicolumn{3}{c}{\textit{Spatial Perception Tasks}} \\
        \midrule
        SoccerNet-GSR~\cite{SoccerNet-GSR} & 42.8K & \makecell[l]{Athlete Detection and Identification \\ Pitch Registration} \\
        \midrule
        SoccerFactory-generated & 2.7M & \makecell[l]{Athlete Detection and Identification \\ Pitch Registration} \\
        \midrule
        \rowcolor{gray!20}
        \multicolumn{3}{c}{\textit{Semantic Reasoning Tasks}} \\
        \midrule
        SoccerNet-v2~\cite{SoccerNetv2} & 1.6M & Event Classification \\
        \midrule
        MatchTime~\cite{rao2024matchtimeautomaticsoccergame} & 7.1K & \makecell[l]{Event Classification \\ Vision-Language Alignment} \\
        \midrule
        SoccerReplay-1988~\cite{rao2025unisoccer} & 3.1M & \makecell[l]{Event Classification \\ Vision-Language Alignment} \\
        \bottomrule
        \end{tabular}
    }
    \vspace{-12pt}
\label{tab:dataset_composition}
\end{table}


\subsection{Visual Encoder}
\label{subsec:architecture}
Our visual encoder~($\Phi_{\mathrm{enc}}$) inherits the ViT~\cite{ViT} structure, consisting of $(L_s + L_{st})$ transformer encoder layers, incorporates a TimeSformer-like~\cite{bertasius2021TimeSformer} design to enable spatiotemporal attention and capture temporal features from videos. 
To balance performance and efficiency, we use spatial attention in the first $L_s$ layers and apply spatiotemporal attention only in the final $L_{st}$ layers. 
The visual feature extraction process comprises three stages: token embedding, spatial encoding, and spatiotemporal encoding, as detailed below.

\vspace{2pt}
\noindent \textbf{Token embedding.}
Given an input video segment ($\mathcal{V} = \{\mathcal{I}_1, \dots, \mathcal{I}_T\} \in \mathbb{R}^{T \times 3 \times H \times W}$), we process each frame following the tokenization procedure of ViT~\cite{ViT}. 
Specifically, each frame is partitioned into $N$ non-overlapping patches, which are then linearly projected into $d$-dimensional embeddings and augmented with learnable spatial positional embeddings $\mathbf{P}_{\mathrm{spa}} \in \mathbb{R}^{N \times d}$. This yields the initial token sequence $\mathbf{z}^{(0)} \in \mathbb{R}^{T \times N \times d}$ across all frames, which preserves spatial layout information within each frame.

\vspace{2pt}
\noindent \textbf{Spatial encoding.}
The token sequence~($\mathbf{z}^{(0)}$) first passes through $L_s$ spatial transformer blocks, which process individual frames separately. 
Each block adopts the standard transformer architecture~\cite{ViT}, consisting of multi-head self-attention, LayerNorm, and feedforward network~(FFN). 
For clarity, we focus on describing the attention mechanism, which is the key modification in our design.
Given a token~($\mathbf{z}_{t,i}^{(l)}$) at spatial position $i$ and temporal position $t$ in the $l$-th block, the spatial self-attention can be expressed as:
\begin{align}
    \mathbf{z}_{t,i}^{(l+1)} = \mathrm{SpatialAttn}(\mathbf{z}_{t,i}^{(l)}, \{\mathbf{z}_{t,j}^{(l)}\}_{j=1}^N)
\end{align}
Here, each token attends exclusively to all tokens at the same temporal position~($t$), with no information exchange across different frames, effectively acting as standard attentions within frames.
The output from the $L_s$-th spatial attention block serves as the extracted \textbf{spatial features}, denoted as $\mathcal{F}_{\mathrm{spa}} = \mathbf{z}^{(L_s)} \in \mathbb{R}^{T \times h \times w \times d}$, which preserve fine-grained spatial details within all frames.

\vspace{2pt}
\noindent \textbf{Spatiotemporal encoding.}
Prior to the spatiotemporal attention blocks, we incorporate learnable temporal positional embeddings~($\mathbf{P}_{\mathrm{tem}} \in \mathbb{R}^{T \times d}$) to the spatial features~($\mathcal{F}_{\mathrm{spa}}$) to encode temporal ordering information.
We adopt a TimeSformer-like~\cite{bertasius2021TimeSformer} approach to extend the remaining $L_{st}$ attention blocks to spatiotemporal attention, where each block alternately performs temporal attention and spatial attention. 
For a token at temporal position $t$ and spatial position $i$, temporal attention can be expressed as:
\begin{align}
    \mathbf{z}_{t,i}^{(l+\frac{1}{2})} = \mathrm{TemporalAttn}(\mathbf{z}_{t,i}^{(l)}, \{\mathbf{z}_{t',i}^{(l)}\}_{t'=1}^T)
\end{align}
Here, each token interacts with all tokens at the same spatial position~($i$), without information exchange across distinct spatial positions. 
Subsequently, a spatial attention layer is applied, which operates in the same manner as previously mentioned, represented as:
\begin{align}
    \mathbf{z}_{t,i}^{(l+1)} = \mathrm{SpatialAttn}(\mathbf{z}_{t,i}^{(l+\frac{1}{2})}, \{\mathbf{z}_{t,j}^{(l+\frac{1}{2})}\}_{j=1}^N)
\end{align}
After $L_s$ spatial and $L_{st}$ spatiotemporal attention blocks, we apply an MAP head~(attention pooling)~\cite{zhai2022scaling} across spatial dimensions to the final features~($\mathbf{z}^{(L_{s} + L_{st})}$) following SigLIP~2~\cite{tschannen2025siglipv2}, and obtain the final \textbf{semantic features}, denoted as $\mathcal{F}_{\mathrm{sem}} = \mathrm{MAP}(\mathbf{z}^{(L_{s} + L_{st})}) \in \mathbb{R}^{T \times d}$, that capture global dynamic semantic information.

This hierarchical design allows the model to capture both spatial details and temporal dynamics in soccer videos, yielding unified representations for downstream tasks.

\subsection{Supervised Multi-task Pretraining}
\label{subsec:multitask_pretraining}

We pretrain our visual encoder with a multi-task framework,
on spatial perception tasks~({\em e.g.}, athlete detection and identification, pitch registration) and semantic reasoning tasks~({\em e.g.}, event classification, vision-language alignment), with lightweight output heads~($\Psi_{\mathrm{out}} = \{ \Psi_{\mathrm{d}}, \Psi_{\mathrm{k}}, \Psi_{\mathrm{l}}, \Psi_{\mathrm{e}}, \Psi_{\mathrm{a}} \}$), detailed as follows.

\vspace{2pt}
\noindent \textbf{Athlete detection and identification.}
The athlete detection and identification head~($\Psi_{\mathrm{d}}$) employs a lightweight Deformable DETR decoder-like~\cite{Deformable-DETR} structure to perform attention between learnable queries and the extracted spatial features~($\mathcal{F}_{\mathrm{spa}}$) to obtain object-level features, which are then passed through three linear layers that predict the athlete bounding box~($\hat{\mathbf{b}}_i$), player role~($\hat{\mathbf{r}}_i$), and jersey number~($\hat{\mathbf{n}}_i$), respectively, expressed as:
\begin{align}
    \{\hat{\mathbf{b}}_i; \hat{\mathbf{r}}_i; \hat{\mathbf{n}}_i\} = \Psi_{\mathrm{d}}(\mathcal{F}_{\mathrm{spa}})
\end{align}
We adopt bounding box regression loss~($\mathcal{L}_{\mathrm{bbox}}$) and classification loss~($\mathcal{L}_{\mathrm{cls}}$) commonly used in detection tasks, while role classification~($\mathcal{L}_{\mathrm{r}}$) and jersey number recognition~($\mathcal{L}_{\mathrm{j}}$) employ focal loss~\cite{FocalLoss}.
The overall objective function for this task is a weighted sum of the above items expressed as:
\begin{align}
    \mathcal{L}_{\mathrm{a}} & = 
    \lambda_{\mathrm{cls}}\mathcal{L}_{\mathrm{cls}} + \lambda_{\mathrm{bbox}}\mathcal{L}_{\mathrm{bbox}} + \lambda_{\mathrm{r}}\mathcal{L}_{\mathrm{r}} + \lambda_{\mathrm{j}}\mathcal{L}_{\mathrm{j}} 
\end{align}
%

\noindent \textbf{Pitch registration.} 
The pitch registration heads~($\{\Psi_{\mathrm{k}}, \Psi_{\mathrm{l}}\}$) inherit the heatmap-based method in PnLCalib~\cite{gutierrez2024pnlcalib}, taking the spatial feature~($\mathcal{F}_{\mathrm{spa}}$) as input and using two convolutional networks~(CNNs) to predict keypoint and endpoint heatmaps for each frame, expressed as:
\begin{align}
    \{\hat{\mathcal{K}}; \hat{\mathcal{L}}\} = \{ \Psi_{\mathrm{k}}(\mathcal{F}_{\mathrm{spa}}); \Psi_{\mathrm{l}}(\mathcal{F}_{\mathrm{spa}}) \}
\end{align}
Specifically, each head uses convolutional layers containing PixelShuffle~\cite{shi2016real} to perform progressive upsampling on features, ultimately using MSE losses between predicted keypoints~($\hat{\mathcal{K}}$) and lines~($\hat{\mathcal{L}}$) and their ground truth as the optimization objective, denoted as $\mathcal{L}_{\mathrm{k}}$ and $\mathcal{L}_{\mathrm{l}}$.

\vspace{2pt}
\noindent \textbf{Event classification.} 
Following UniSoccer~\cite{rao2025unisoccer}, the event classification head~($\Psi_{\mathrm{e}}$) adopts a two-layer transformer encoder, followed by temporal average pooling and a linear classifier, which predicts soccer event~($\hat{\mathbf{e}}$) based on semantic reasoning features~($\mathcal{F}_{\mathrm{sem}}$), expressed as:
\begin{align}
    \hat{\mathbf{e}} = \Psi_{\mathrm{e}}(\mathcal{F}_{\mathrm{sem}})
\end{align}
The head is optimized using cross-entropy loss~($\mathcal{L}_{\mathrm{e}}$).

\vspace{2pt}
\noindent \textbf{Vision-language alignment.} 
Similar to UniSoccer~\cite{rao2025unisoccer}, 
this alignment head~($\Psi_{\mathrm{a}}$) first performs temporal average pooling on semantic reasoning features~($\mathcal{F}_{\mathrm{sem}}$) to obtain video representations and computes similarity~(${\mathbf{s}}$) with text embeddings of commentary~($\mathcal{T}$) extracted by an off-the-shelf SigLIP~2 text encoder~($\Phi_{\mathrm{text}}$)~\cite{tschannen2025siglipv2}, expressed as:
\begin{align}
    \mathbf{s} = \Psi_{\mathrm{a}}(\mathcal{F}_{\mathrm{sem}}, \Phi_{\mathrm{text}}(\mathcal{T}))
\end{align}
The head is optimized using SigLIP loss~\cite{zhai2023SigLIP} in a contrastive learning manner, denoted as $\mathcal{L}_{\mathrm{con}}$.

\vspace{2pt}
\noindent \textbf{Overall training loss.} 
The final training objective~($\mathcal{L}_{\mathrm{total}}$) combines all task-specific ones with corresponding weights:
\begin{align}
    \mathcal{L}_{\mathrm{total}} & = \lambda_{\mathrm{a}}\mathcal{L}_{\mathrm{a}} + \lambda_{\mathrm{k}}\mathcal{L}_{\mathrm{k}} + \lambda_{\mathrm{l}}\mathcal{L}_{\mathrm{l}} + \lambda_{\mathrm{e}}\mathcal{L}_{\mathrm{e}} + \lambda_{\mathrm{con}}\mathcal{L}_{\mathrm{con}}
\end{align}
This unified multi-task pretraining enables SoccerMaster to learn complementary spatial and semantic representations.

\begin{table*}[tphb]
    \caption{
      \textbf{Performance Comparison on Pretraining Tasks.}
      We compare SoccerMaster against both general-purpose VFMs (SigLIP2-L/16-512~\cite{tschannen2025siglipv2} and DINOv3-L/16~\cite{simeoni2025dinov3}) and the soccer-specific MatchVision~\cite{rao2025unisoccer} using frozen encoders with trainable task-specific heads. ``SoccerFactory Data'' indicates training augmented with automatically generated data from our pipeline alongside existing datasets. 
      Metrics include AP@50 and mAP for detection, jersey number~(jn) and role classification accuracy, keypoint/line detection metrics, event classification accuracy, and video-commentary retrieval top-1 accuracy~(computed within batches of 48). 
    }
    \vspace{-3pt}
    \footnotesize
    \centering
    \renewcommand{\arraystretch}{1.1}
    \resizebox{0.99\linewidth}{!}{
        \begin{tabular}{l | c | cccc | ccc | ccc | c | c }
        \toprule
        \multirow{2}{*}{\textbf{Encoder}} & \multirow{2}{*}{\textbf{SoccerFactory Data}} & \multicolumn{4}{c|}{\textbf{Athlete Detection and Identification}} & \multicolumn{3}{c|}{\textbf{Keypoints Detection}} & \multicolumn{3}{c|}{\textbf{Lines Detection}} & \textbf{Event Cls.} & \textbf{Alignment} \\
        \cline{3-14}
         & & AP@50 & mAP & jn & role & accuracy & precision & recall & accuracy & precision & recall & accuracy & top-1 \\
        \midrule
        SigLIP 2 & & 69.1 & 29.2 & \underline{78.8} & 97.5 & 92.5 & 80.0 & 82.0 & 90.0 & 80.2 & 85.9 & 49.8 & 3.4 \\
        SigLIP 2 & \Checkmark & \underline{72.3} & \underline{32.0} & 78.2 & 97.3 & 93.7 & 80.0 & \underline{83.4} & \underline{93.6} & 84.7 & \underline{92.3} & - & - \\
        DINOv3 & & 61.8 & 22.9 & 75.6 & \underline{98.5} & 92.5 & 77.2 & 80.2 & 91.0 & 82.0 & 87.1 & 51.8 & - \\
        DINOv3 & \Checkmark & 70.2 & 28.0 & 76.1 & 98.1 & \underline{94.0} & 81.4 & \underline{83.4} & 92.8 & 82.7 & 92.7 & - & - \\
        MatchVision & \Checkmark & 51.9 & 17.0 & 74.9 & 94.1 & 92.0 & \underline{85.2} & 69.8 & 90.3 & \underline{86.6} & 81.5 & \underline{65.3} & 4.0 \\
        \midrule
        \rowcolor{gray!20}
        \textbf{Ours} & \Checkmark & \textbf{91.5} & \textbf{49.5} & \textbf{79.7} & \textbf{99.1} & \textbf{95.3} & \textbf{85.9} & \textbf{84.1} & \textbf{95.7} & \textbf{91.4} & \textbf{92.4} & \textbf{77.2} & \textbf{39.0} \\
        
        \bottomrule
        \end{tabular}
    }
    \vspace{-3pt}
    \label{tab:comparison}
\end{table*}

\subsection{Downstream Task Adaptation}
\label{subsec:downstream_tasks}
After pretraining, our \textbf{SoccerMaster} can then be seamlessly adapted to various downstream tasks. 
By introducing task-specific heads, we demonstrate the versatility and generalization ability of our learned representations, covering commentary generation, camera calibration, and multiple object tracking, as detailed below.

\vspace{2pt}
\noindent \textbf{Commentary generation.}
Similar to~\cite{rao2025unisoccer, rao2024matchtimeautomaticsoccergame}, we adopt a Q-Former~\cite{blip2} architecture to perform temporal aggregation on semantic features~($\mathcal{F}_{\mathrm{sem}}$) and employ a linear layer to project the pooled tokens into the embedding space of a large language model~(Llama-3-8B~\cite{llama3}), which serve as prefix embeddings for auto-regressive text generation.

\vspace{2pt}
\noindent \textbf{Camera calibration.}
Building upon the field keypoints and lines detected by our pitch registration heads~($\{\Psi_{\mathrm{k}}, \Psi_{\mathrm{l}}\}$), we apply the Point and Line (PnL) refinement module~\cite{gutierrez2024pnlcalib} to estimate the camera intrinsic and extrinsic parameters.

\vspace{2pt}
\noindent \textbf{Multiple object tracking.}
Following MOTIP~\cite{motip}, we perform tracking by formulating object association as a classification task.
Concretely, we first extract object-level features from the DETR decoder of the athlete detection and identification head~($\Psi_{\mathrm{d}}$). 
Then, we construct historical trajectories by concatenating these features with learnable ID embeddings from an ID dictionary, which serves as in-context identity prompts. 
Finally, an ID decoder, implemented as a transformer decoder, predicts consistent identity labels for each detected athlete in the current frame by attending to historical trajectories from previous frames.
\section{Experiments}
\label{sec:experiments}
This section presents comprehensive evaluations of \textbf{SoccerMaster} across various downstream tasks. 
Concretely, we begin with implementation details in Sec.~\ref{subsec:implementation_details}, followed by the comparisons on downstream tasks in Sec.~\ref{subsec:downstream_tasks_performance}, and conduct ablation studies in Sec.~\ref{subsec:ablation_studies}. 

\subsection{Implementation Details}
\label{subsec:implementation_details}
\textbf{SoccerMaster} model adopts a hierarchical vision transformer architecture, initialized with weights from siglip2-large-patch16-512~\cite{tschannen2025siglipv2}. 
The visual encoder consists of $L_s = 16$ spatial transformer blocks followed by $L_{st} = 8$ spatiotemporal transformer blocks, with a hidden dimension of $d = 1024$. 
Input videos are uniformly sampled into clips of $T=30$ frames and processed at a resolution of $512 \times 512$ with a patch size of $16 \times 16$. 
Further implementation details are provided in Sec.~\ref{sec:more_implementation_details} of the \textbf{Appendix}.

\subsection{Performance Comparison}
\label{subsec:downstream_tasks_performance}
We conduct comprehensive evaluations across multiple tasks to validate the effectiveness of our model.

\vspace{2pt}
\noindent \textbf{Comparison on pretraining tasks.} 
We compare against general-purpose baselines~(SigLIP2~\cite{tschannen2025siglipv2} and DINOv3~\cite{simeoni2025dinov3}) and the soccer-specific vision model~(MatchVision) on the four core pretraining tasks.
Dataset details are provided in Tab.~\ref{tab:dataset_composition}.
To ensure a fair comparison, we adopt the following evaluation protocol:
for baseline models~(SigLIP2, DINOv3, and MatchVision), we freeze their pretrained encoders and train only lightweight task-specific heads on our pretraining datasets; 
While for \textbf{SoccerMaster}, as it has natively pretrained on these tasks, we directly evaluate it without further tuning. 
This setup rigorously assesses the representation quality of each encoder by maintaining consistent task-specific head architectures.

As shown in Tab.~\ref{tab:comparison}, we can derive the following observations: 
(i) \textbf{SoccerMaster} achieves substantial improvements over the second-best methods, notably improving athlete detection and identification by \textbf{+17.5\%} mAP and event classification by \textbf{+11.9\%} accuracy;
(ii) our model exhibits remarkable performance in the vision-language alignment~(retrieval) task, achieving a \textbf{39.0\%} top-1 accuracy that largely surpasses all baselines. This gap highlights specific weaknesses in baselines: general-purpose models like SigLIP~2 suffer from severe domain gaps~(only 3.4\% accuracy), while DINOv3 lacks a corresponding text encoder for alignment evaluation; 
and
(iii) although tailored for soccer data, MatchVision still lacks comprehensive task coverage.
While it demonstrates competitive performance in keypoint detection, line detection, and event classification, it struggles significantly in both athlete detection and identification and vision-language alignment. 
This may be attributed to its over-reliance on the joint optimization of vision and language encoders, which neglects fine-grained spatial perception tasks and leads to misalignment with the pretrained SigLIP~2 text encoder.

\begin{table}[t]
    \centering
    \renewcommand{\arraystretch}{1.1} 
    \caption{
        \textbf{Comparison on Camera Calibration.} 
        Here, JaC$_\gamma$, CR, and FS denote the Jaccard index at threshold $\gamma$, completeness rate, and final score, respectively.
        For our model, * denotes zero-shot inference and $^\dagger$ indicates fine-tuned training.
    }
    \vspace{-3pt}
    \footnotesize
    \resizebox{0.98\linewidth}{!}{
    \begin{tabular}{l|l|ccc|c|c|c}
        \toprule
        
        \multirow{2}{*}{\centering \makecell{\textbf{Dataset}}} & 
        \multirow{2}{*}{\centering \makecell{\textbf{Method}}} & 
        \multicolumn{3}{c|}{\textbf{JaC$_\gamma$ [\%]}} & 
        \multirow{2}{*}{\centering \makecell{\textbf{CR}}} & 
        \multirow{2}{*}{\centering \makecell{\textbf{FS}}} & 
        \multirow{2}{*}{\centering \makecell{\textbf{Resolution}}} \\
        
        \cline{3-5}
        
         & & 5 & 10 & 20 & & & \\
        \midrule
        
        \multirow{4}{*}{\centering \makecell{SN22\\-test\\-center}} 
         & PnlCalib & \textbf{80.3} & 91.9 & 94.2 & 97.9 & \textbf{78.6} & 960$\times$540 \\
         & PnlCalib & 70.2 & 86.8 & 91.7 & 96.4 & 67.6 & 512$\times$512 \\
         & \textbf{Ours}* & 70.5 & \underline{92.0} & \textbf{94.8} & \textbf{99.4} & 70.1 & 512$\times$512 \\
         & \textbf{Ours}$^\dagger$ & \underline{76.9} & \textbf{92.3} & \underline{94.4} & \underline{98.6} & \underline{75.8} & 512$\times$512 \\
        \midrule
        
        \multirow{4}{*}{\centering \makecell{SN23\\-test}}
         & PnlCalib & \textbf{76.7} & \textbf{87.2} & \textbf{90.1} & \textbf{79.5} & \textbf{60.9} & 960$\times$540 \\
         & PnlCalib & 66.9 & 82.4 & 86.7 & 78.4 & 51.8 & 512$\times$512 \\
         & \textbf{Ours}* & 59.5 & 78.1 & 81.6 & 75.2 & 44.8 & 512$\times$512 \\
         & \textbf{Ours}$^\dagger$ & \underline{71.1} & \underline{85.4} & \underline{87.8} & \underline{79.1} & \underline{56.2} & 512$\times$512 \\
        \bottomrule
    \end{tabular}
    }
    \label{tab:performance_camera_calibration}
    \vspace{-12pt}
\end{table}

\begin{table}[htbp]
    \caption{
        \textbf{Comparison on Multiple Object Tracking.} 
        Notably, our model is the only one employing an end-to-end setting.
    }
    \vspace{-3pt}
    \centering
    \footnotesize
    \resizebox{0.96\linewidth}{!}{
    \begin{tabular}{l | c c c c c }
    \toprule
    \textbf{Method} & \textbf{HOTA} & \textbf{DetA} & \textbf{AssA} & \textbf{MOTA} & \textbf{IDF1} \\
    \midrule
     YOLOv8+OC-SORT~\cite{cao2023observation} & 54.6 & \underline{63.5} & 47.1 & \underline{76.2} & 62.5 \\
     YOLOv8+StrongSort++~\cite{du2023strongsort} & 56.2 & 62.9 & 50.3 & 75.0 & 66.5 \\
     YOLOv8+PRTreID~\cite{prtreid} & \textbf{59.8} & 61.1 & \textbf{58.6} & 73.1 & \underline{74.5} \\
     \midrule
     \textbf{Ours}+MOTIP~\cite{motip} & \underline{59.1} & \textbf{65.2} & \underline{53.9} & \textbf{81.6} & \textbf{74.6} \\
    \bottomrule
    \end{tabular}
    }
    \label{tab:performance_tracking}
    \vspace{-6pt}
    
\end{table}

\begin{table}[htbp]
    \caption{
        \textbf{Comparison on Commentary Generation.}
    }
    \vspace{-3pt}
    \centering
    \footnotesize
    \setlength{\tabcolsep}{0.10cm} 
    \resizebox{0.96\columnwidth}{!}{
        \begin{tabular}{l | ccccc}
        \toprule
        \textbf{Visual Encoder} & \textbf{BLEU@1} & \textbf{BLEU@4} & \textbf{METEOR} & \textbf{ROUGE-L} & \textbf{CIDEr} \\
        \midrule
        SigLIP 2 & 28.9 & 6.9 & 24.7 & 25.4 & 28.3 \\
        MatchVision & \underline{30.9} & \underline{8.7} & \textbf{26.9} & \textbf{27.6} & \underline{35.7} \\
        \midrule
        \textbf{Ours} & \textbf{31.3} & \textbf{8.9} & \underline{26.2} & \underline{26.6} & \textbf{38.6} \\
        \bottomrule
        \end{tabular}
    }
    \vspace{-6pt}
    \label{tab:performance_commentary}
\end{table}

\vspace{2pt}
\noindent \textbf{Comparison on camera calibration.} 
We evaluate camera calibration performance against the state-of-the-art method, namely PnlCalib~\cite{gutierrez2024pnlcalib}, on the SoccerNet-22 and SoccerNet-23 benchmarks.
As detailed in Tab.~\ref{tab:performance_camera_calibration}, we adhere to the evaluation protocol in previous works~\cite{gutierrez2024pnlcalib, SoccerNet2022, CameraCalibration}, and report the Jaccard index~($\mathrm{JaC}_\gamma$) at thresholds of 5, 10, and 20 pixels, along with the completion rate~(CR) and final score~(FS).
Specifically, the Jaccard index~($\mathrm{JaC}_\gamma$) measures calibration accuracy based on reprojection error; it is calculated as TP/(TP+FN+FP), where a pitch line is deemed correctly detected only if all its points exhibit reprojection errors below the threshold $\gamma$.
The completion rate~($\mathrm{CR}$) quantifies the proportion of images for which the method successfully produces camera parameters, while the final score, defined as $\mathrm{FS} = \mathrm{CR} \times \mathrm{JaC}_5$, serves as the primary evaluation criterion.
Notably, PnlCalib is originally optimized for a higher resolution of 960$\times$540, achieving state-of-the-art performance in that setting~(78.6 FS on SN22-test-center and 60.9 FS on SN23-test).
However, since SoccerMaster operates at a 512$\times$512 input resolution, we evaluate all methods at this resolution to ensure a fair comparison.

The results reveal two observations: 
(i) on zero-shot evaluation, SoccerMaster already outperforms PnlCalib on SN22-test-center~(70.1 vs.~67.6 FS). On the more challenging SN23-test, which contains diverse non-main camera views, our zero-shot model achieves 44.8 FS compared to PnlCalib's 51.8 FS;
(ii) once fine-tuned, SoccerMaster establishes a new standard at this resolution, surpassing PnlCalib by \textbf{+8.2} FS on SN22~(75.8 vs.~67.6) and \textbf{+4.4} FS on SN23 (56.2 vs.~51.8).
These results demonstrate that SoccerMaster has learned strong and transferable representations for field geometry, effectively generalizing across diverse camera views and challenging scenarios.

\vspace{2pt}
\noindent \textbf{Comparison on multiple object tracking.} 
As shown in Tab.~\ref{tab:performance_tracking}, we evaluate our method on the SoccerNet-tracking benchmark~\cite{SoccerNetv3-Tracking} using standard MOT metrics~(HOTA~\cite{HOTA}, DetA~\cite{HOTA}, AssA~\cite{HOTA}, MOTA~\cite{MOTA} and IDF1~\cite{IDF1}).
Unlike the tracking-by-detection paradigm relying on complex multi-stage components such as separate detectors, Re-ID models, and post-processing modules, SoccerMaster employs a much simpler end-to-end pipeline while achieving competitive results.
Concretely, our model exhibits strong detection capability, securing the best DetA~(65.2) and the second-best performance on association AssA~(53.9). Overall, our model demonstrates competitive results across all key metrics: HOTA~(59.1), MOTA~(81.6), and IDF1~(74.6). These results highlight the strong transferability of our learned representations, which can be effectively adapted to tracking tasks through lightweight fine-tuning, significantly simplifying the inference pipeline.

\vspace{2pt}
\noindent \textbf{Comparison on commentary generation.} 
We evaluate the commentary generation task on the SN-Caption-test-align benchmark~\cite{rao2024matchtimeautomaticsoccergame}, which comprises 49 soccer matches with precise manual annotations and serves as the only available dataset with accurate temporal alignment.
For fair comparison, all models are fine-tuned on the MatchTime~\cite{rao2024matchtimeautomaticsoccergame} dataset before evaluation.
Tab.~\ref{tab:performance_commentary} presents the results using standard metrics of natural language generation, including BLEU~\cite{bleu}, METEOR~\cite{meteor}, ROUGE-L~\cite{rouge}, and CIDEr~\cite{cider}. 
Our model achieves the best results on BLEU@1~(31.3), BLEU@4~(8.9), and CIDEr~(38.6), with notable improvements over MatchVision~(\textbf{+0.4} on BLEU@1 and \textbf{+2.9} on CIDEr). While our scores on METEOR~(26.2) and ROUGE-L~(26.6) are slightly below MatchVision, they remain competitive and substantially outperform the general-purpose SigLIP2~\cite{tschannen2025siglipv2} baseline.
The strong performance on CIDEr, which measures semantic similarity and consensus with human references, demonstrates that our multi-task pretraining strategy fosters rich semantic representations that effectively transfer to language generation tasks.

\begin{table*}[!t]
    \caption{
        \textbf{Ablation Study on the Impact of Automatically Generated Spatial Annotations.}
    }
    \vspace{-3pt}
    \centering
    \footnotesize
    \renewcommand{\arraystretch}{1.1}
    \resizebox{\linewidth}{!}{
        \begin{tabular}{c | cccc | ccc | ccc | c | c}
        \toprule
        \multirow{2}{*}{\makecell{\textbf{SoccerFactory} \\ \textbf{Data}}} & \multicolumn{4}{c|}{\textbf{Athlete Detection and Identification}} & \multicolumn{3}{c|}{\textbf{Keypoints Detection}} & \multicolumn{3}{c|}{\textbf{Lines Detection}} & \textbf{Event Classification} & \textbf{Alignment} \\
        \cline{2-13}
         & AP@50 & mAP & jn & role & accuracy & precision & recall & accuracy & precision & recall & accuracy & top-1 \\
        \midrule
         & 77.7 & 30.2 & 75.0 & 97.3 & 92.9 & 89.6 & 70.5 & 93.4 & 93.9 & \textbf{86.0} & \textbf{71.6} & 35.0 \\
        \Checkmark & \textbf{82.0} & \textbf{37.5} & \textbf{76.5} & \textbf{98.1} & \textbf{93.1} & \textbf{90.1} & \textbf{70.7} & \textbf{93.7} & \textbf{96.1} & 85.2 & 70.5 & \textbf{36.8} \\
        \bottomrule
        \end{tabular}
    }
    \label{tab:ablation_pesudo_data}
    \vspace{-6pt}
\end{table*}

\subsection{Ablation Study}
\label{subsec:ablation_studies}
Tab.~\ref{tab:ablation_pesudo_data} shows the effectiveness of incorporating automatically generated spatial annotations for pretraining. 
Concretely, to balance computational efficiency, we utilize a compact variant of SoccerMaster, which is configured with a $224\times 224$ input resolution, $16\times 16$ patch size, $L_s = 8$, and $L_{st} = 4$ encoding layers.
We observe that the inclusion of pipeline-generated data yields substantial improvements in spatial perception tasks, particularly in athlete detection, achieving \textbf{+4.3} AP@50~(82.0 vs.~77.7) and \textbf{+7.3} mAP~(37.5 vs.~30.2).
Meanwhile, other pretraining tasks, including pitch registration, event classification, and vision-language alignment, maintain stable or slightly improved performance.
This confirms that our automatically curated annotations provide a reliable and scalable source of training data.
More ablation studies and qualitative results can be found in Sec.~\ref{sec:ablation_on_multi-task_pretraining} and Sec.~\ref{sec:qualitative_results} of the \textbf{Appendix}.

\section{Conclusion}
\label{sec:conclusion}
This paper aims to advance soccer visual understanding by proposing a unified framework that consolidates diverse tasks into a single model.
Concretely, we present \textbf{SoccerMaster}, the first soccer-specific vision foundation model capable of handling comprehensive tasks spanning fine-grained spatial perception and high-level semantic reasoning through supervised multi-task pretraining.
To facilitate large-scale pretraining, we introduce \textbf{SoccerFactory}, an automated data curation pipeline that generates scalable spatial annotations from broadcast videos. Integrating these annotations with multiple existing soccer datasets, we construct a comprehensive pretraining dataset.
Extensive evaluations demonstrate that SoccerMaster serves as an effective soccer-specific vision foundation model, where simple fine-tuning yields state-of-the-art performance across diverse downstream tasks.
We believe this work establishes a new paradigm for soccer visual understanding by offering a unified, adaptable framework that reduces the engineering overhead of developing task-specific systems and provides a solid foundation for future advances in soccer AI.

\section*{Acknowledgments}
Weidi would like to acknowledge the funding from Scientific Research Innovation Capability Support Project for Young Faculty~(ZY-GXQNJSKYCXNLZCXM-I22).

\clearpage

{
    \small
    \bibliographystyle{ieeenat_fullname}
    \bibliography{main}
}

\onecolumn
{
    \centering
    \Large
    \textbf{SoccerMaster: A Vision Foundation Model for Soccer Understanding}  \\
    \vspace{0.5em} Supplementary Material \\
    \vspace{1.0em}
}

\appendix
{
  \hypersetup{linkcolor=black}
  \tableofcontents
}

\clearpage

\section{Pretraining Dataset Construction and Integration}
\label{sec:soccerfactory_dataset_construction}
To facilitate comprehensive multi-task pretraining, we construct a unified, large-scale dataset by integrating the automatically curated data from our pipeline, \textbf{SoccerFactory}, with multiple existing soccer video datasets~\cite{SoccerNet-GSR, SoccerNetv2, rao2024matchtimeautomaticsoccergame, rao2025unisoccer}.
This section details the datasets used for each pretraining task, their preprocessing strategies, and integration into our unified training framework. 
Specifically, we first detail the data automatically curated by SoccerFactory in Sec.~\ref{subsec:gsr_format}, then introduce the integration of spatial and semantic tasks in Sec.~\ref{spatialtask} and Sec.~\ref{semantictask}, along with the statistics of the integrated dataset in Sec.~\ref{datastatistics}.

\subsection{Details of SoccerFactory-Generated Data}
\label{subsec:gsr_format}
Our proposed automated curation pipeline, \textbf{SoccerFactory}, generates high-quality, frame-level annotations for spatial perception tasks on soccer broadcast videos.
Specifically, each video clip is accompanied by a \texttt{.json} file containing dense per-frame annotations of athletes, field keypoints, and lines, structured as follows:

\vspace{6pt}
\noindent \textbf{Frame-level annotations.} 
Each frame in the video is associated with holistic spatial annotations, including athlete detection, pitch registration, and role classification information:
{
\small
\begin{verbatim}
{
  "athletes": [...],                         # List of athlete annotations
  "keypoints": [...],                        # Pitch keypoints
  "lines": [...],                            # Pitch line segments
  "K": [...],                                # Camera intrinsic matrix
  "Rt": [...],                               # Camera extrinsic matrix
  "valid_cam_params": True                   # Calibration validity flag
}
\end{verbatim}
}

\vspace{2pt} 
\noindent \textbf{Athlete annotations.} 
For each athlete detected in a video frame, we provide the attributes describing this player, including bounding box coordinates, track ID, jersey number, legibility score, role, and team affiliation, for example:
{
\small
\begin{verbatim}
{
  "bbox_ltwh": [1116.5, 679.5, 50.8, 98.2],  # Bounding box (left, top, width, height)
  "track_id": 4,                             # Tracklet unique identity
  "jersey_number": "10",                     # Jersey number (tracklet-level)
  "legibility_score": 0.67,                  # Jersey number legibility score
  "role": "player",                          # Player role
  "team": "right"                            # Team affiliation
}
\end{verbatim}
}

\vspace{2pt} 
\noindent \textbf{Field keypoint annotations.} 
We provide 2D keypoints detected by the field keypoint detector, indexed according to the semantic definitions in PnlCalib~\cite{gutierrez2024pnlcalib}.
Here is an example:
{
\small
\begin{verbatim}
{
  2: {"x": 984.0, "y": 348.0, "p": 0.800},   # Coordinates and confidence
  32: {"x": 984.0, "y": 460.0, "p": 0.846},  
  ...                                        
}
\end{verbatim}
}

\vspace{2pt} 
\noindent \textbf{Field line annotations.} 
We provide detected field line segments as sequences of 2D points in normalized image coordinates. Each line is identified by its semantic label following SoccerNet-v3~\cite{SoccerNetv3}. 
An example is as follows:
{
\small
\begin{verbatim}
{
  "Circle central": [                        # Sequences of points
    {"x": 0.513, "y": 0.426}, 
    {"x": 0.388, "y": 0.441},
    {"x": 0.329, "y": 0.470}, ...
  ],
  "Middle line": [
    {"x": 0.513, "y": 0.322}, 
    {"x": 0.513, "y": 0.426},
    {"x": 0.515, "y": 0.485}, ...
  ],
  ...                                        
}
\end{verbatim}
}

\subsection{Spatial Perception Tasks}
\label{spatialtask}
During pretraining, the spatial perception module addresses two primary tasks: (i) athlete detection and identification, and (ii) pitch registration. These tasks leverage annotated data from two sources: automatically generated data from our pipeline and existing data from SoccerNet-GSR~\cite{SoccerNet-GSR}.

\subsubsection{Data Sources Introduction}
\noindent \textbf{SoccerFactory-generated data} are produced by our automated data curation pipeline, with annotation format details provided in Sec.~\ref{subsec:gsr_format}. 
We extract main-camera clips from 500 matches in the SoccerNetv2~\cite{SoccerNetv2} dataset by utilizing the ground truth from the SoccerNet Camera Shot Segmentation task~\cite{SoccerNetv2}, specifically selecting segments labeled as ``main camera center'' to exclude replays, close-ups, and alternative angles. To align with SoccerNet-GSR~\cite{SoccerNet-GSR}, these extracted clips are further partitioned into segments with a maximum duration of 30 seconds.

\vspace{2pt} 
\noindent \textbf{SoccerNet-GSR}~\cite{SoccerNet-GSR} comprises 200 uncut, 30-second broadcast video clips captured by a single moving camera, providing realistic and challenging scenarios for spatial understanding tasks.
It features extensive annotations, including over 9.37 million line points for pitch localization and camera calibration, as well as over 2.36 million athlete positions on the pitch labels with their roles, teams, and jersey numbers. 

\subsubsection{Datasets Integration Details}
\noindent \textbf{Data preprocessing and integration.} 
For both pipeline-generated data and SoccerNet-GSR, we retain a unified subset of annotations per frame used in pretraining. 
Specifically, each frame contains the following information:
{
\small
\begin{verbatim}
{
  "athletes": [...],                         # List of athlete annotations
  "keypoints": [...],                        # Pitch keypoints
  "lines": [...]                             # Pitch line segments
}
\end{verbatim}
}

\vspace{2pt} 
\noindent For each athlete detection, we maintain the following structured data, for example:
{
\small
\begin{verbatim}
{
  "bbox_ltwh": [1116.5, 679.5, 50.8, 98.2],  # Bounding box
  "track_id": 4,                             # Track identity
  "jersey_number": "10",                     # Jersey number (bbox-level)
  "role": "player"                           # Player role
}
\end{verbatim}
}

\vspace{2pt} 
\noindent \textbf{Jersey number filtering.} 
Since SoccerNet-GSR provides tracklet-level jersey number annotations, we employ a legibility classifier~\cite{JerseyNumber} to compute frame-level legibility scores for each detection. 
For both datasets, jersey numbers with legibility scores below 0.5 are set to \texttt{null}.
This aligns with our athlete detection task definition, which specifies that jersey numbers should be marked as \texttt{null} when they are not clearly visible due to occlusion or viewpoint variations.

\vspace{2pt} 
\noindent \textbf{Pitch keypoints and lines generation.} 
For pitch registration annotations, we adopt different strategies for the two datasets based on their characteristics. 
For SoccerNet-GSR, we directly utilize the manually annotated pitch keypoints and line segments provided in the dataset. 
For data produced by our pipeline, we leverage the estimated camera parameters~(intrinsic matrix $\mathbf{K}$ and extrinsic matrix $[\mathbf{R}|\mathbf{t}]$) to project a standard soccer pitch model containing line segments onto the 2D image plane, and then extract the visible portion within the frame boundaries to serve as ground truth annotations. 
This projection-based approach for our curated data provides geometrically consistent and precise annotations at scale, while the manual annotations from SoccerNet-GSR offer high-quality ground truth for validation and fine-grained supervision.

\subsection{Semantic Reasoning Tasks}
\label{semantictask}
In our multi-task pretraining, the semantic reasoning tasks focus on event classification and vision-language alignment.
We utilize data from three datasets: SoccerNet-v2~\cite{SoccerNetv2}, MatchTime~\cite{rao2024matchtimeautomaticsoccergame}, and SoccerReplay-1988~\cite{rao2025unisoccer}, as detailed below.

\subsubsection{Data Sources Introduction}
\noindent \textbf{SoccerNet-v2.} 
This dataset contains over 110k event labels across 500 matches, originally categorized into 17 distinct classes. 
Following UniSoccer~\cite{rao2025unisoccer}, we systematically remap these labels into 24 standardized event categories. 
For example, ``Direct free-kick'' and ``Indirect free-kick'' are merged into a single ``Free Kick'' category, while ``Penalty'' outcomes are distinguished as ``Penalty'' and ``Penalty Missed.''

\vspace{2pt} 
\noindent \textbf{MatchTime.} 
The MatchTime dataset~\cite{rao2024matchtimeautomaticsoccergame} consists of 471 matches, featuring high-quality aligned video-commentary pairs, which is the temporally aligned version of SoccerNet-Caption~\cite{densecap}. 
To leverage this dataset for both event classification and vision-language alignment, we adopt the prompt-based summarization approach from UniSoccer~\cite{rao2025unisoccer} to extract event categories from the commentary text, thereby associating each video clip with a corresponding event label.

\vspace{2pt} 
\noindent \textbf{SoccerReplay-1988.} 
This large-scale dataset provides 1,988 matches with both event labels and temporally aligned commentaries, serving as the primary resource for joint training in event classification and vision-language alignment.

\subsubsection{Datasets Integration Details}
We integrate all samples from these three datasets into a unified training corpus. 
While all three datasets provide event labels for event classification training, only MatchTime and SoccerReplay-1988 include text commentaries. 
Consequently, during training, the event classification loss is computed on all samples, whereas the vision-language alignment loss is calculated exclusively on samples with available commentaries. 
The final training utilizes 30-second soccer video clips alongside their corresponding event labels~(spanning 24 classes) and commentaries when applicable. 
The train/valid/test split settings remain consistent with the original dataset configurations.

\begin{table}[t]
    \caption{
        \textbf{Statistics of the Integrated Pretraining Dataset.} 
        An overview of the train/validation/test sample counts for all datasets employed in multi-task pretraining. 
        Here, data generated by SoccerFactory is exclusively used for training. 
        Meanwhile, SoccerNet-GSR additionally includes samples from 36 sequences for the SoccerNet challenge evaluation.
    }
    \centering
    \footnotesize
    \begin{tabular}{l|c|c|c|l}
    \toprule
    \textbf{Dataset} & \textbf{Train} & \textbf{Valid} & \textbf{Test} & \textbf{Tasks} \\
    \midrule
    \rowcolor{gray!20}
    \multicolumn{5}{c}{\textit{Spatial Perception Tasks~(Dense Sampling, 25 FPS)}} \\
    \midrule
    SoccerNet-GSR~\cite{SoccerNet-GSR} & 1,425 & 1,450 & 1,225 & \makecell[l]{Athlete Detection \\ Pitch Registration} \\
    \midrule
    SoccerFactory-generated & 94,628 & -- & -- & \makecell[l]{Athlete Detection \\ Pitch Registration} \\
    \midrule
    \rowcolor{gray!20}
    \multicolumn{5}{c}{\textit{Semantic Reasoning Tasks~(Sparse Sampling, 1 FPS)}} \\
    \midrule
    SoccerNet-v2~\cite{SoccerNetv2} & 54,448 & 17,491 & 18,641 & Event Classification \\
    \midrule
    MatchTime~\cite{rao2024matchtimeautomaticsoccergame} & 24,027 & 3,144 & 3,256 & \makecell[l]{Event Classification \\ Vision-Language Alignment} \\
    \midrule
    SoccerReplay-1988~\cite{rao2025unisoccer} & 104,080 & 17,892 & 17,402 & \makecell[l]{Event Classification \\ Vision-Language Alignment} \\
    \bottomrule
    \end{tabular}
    \vspace{-6pt}
    \label{tab:dataset_splits}
\end{table}

\subsection{Data Statistics}
\label{datastatistics}
To accommodate the different nature of spatial perception and semantic reasoning tasks, we employ tailored sampling strategies for each category. 
For \textbf{spatial perception tasks}, we sample dense frames from video clips at 25 FPS, where each training sample comprises 30 consecutive frames. 
Conversely, for \textbf{semantic reasoning tasks}, we adopt a sparse sampling strategy at 1 FPS; here, each video clip constitutes a single sample centered around the specific event timestamp or commentary moment. 
Tab.~\ref{tab:dataset_splits} presents the detailed statistics of data samples across all datasets within our multi-task pretraining framework.
The resulting training resource, termed \textbf{SoccerFactory}, aggregates these diverse data to facilitate comprehensive pretraining.

\section{More Implementation Details}
\label{sec:more_implementation_details}

\subsection{Implementation details of SoccerMaster}

\textbf{SoccerMaster} is initialized from SigLIP 2-large-patch16-512~\cite{tschannen2025siglipv2}, with zero-initialized temporal attention layers and temporal positional embeddings to enable temporal modeling while preserving the pretrained spatial features.
The model is optimized using the AdamW~\cite{adamW} optimizer with differentiated learning rates across model components, as detailed in Tab.~\ref{tab:learning_rates}. 
All parameters except temporal positional embeddings are regularized with a weight decay of $1.0 \times 10^{-4}$. 
These specific learning rates account for the distinct initialization states and functional roles of each module.
We train for a total of 20 epochs, with the first epoch using linear warm-up and subsequent epochs following a CosineAnnealingLR~\cite{adam} schedule. 

To stabilize multi-task training, we apply gradient normalization and clipping independently to the backbone and each task-specific head.
By clipping the gradient norm to a maximum of 0.1 for all components, we prevent any single task from dominating the optimization process.
The multi-task learning objective integrates various loss components with carefully tuned weights, as shown in Tab.~\ref{tab:loss_weights}. 
These weights serve to balance optimization signals across tasks characterized by varying sample sizes and loss magnitudes.

\begin{table}[t]
    \centering
    \caption{
    \textbf{Learning Rates for Model Components.} 
        Here, temporal-related components\textsuperscript{*} include temporal attention layers and temporal positional embeddings, which are zero-initialized.
    }
    \label{tab:learning_rates}
    \footnotesize
    \begin{tabular}{l|c|c|c|c|c|c|c}
    \toprule
    \textbf{Component} & \makecell{Backbone \\ (Temporal- \\ agnostic)} & \makecell{Backbone \\ (Temporal- \\ related)\textsuperscript{*}} & \makecell{Athlete \\ Detection \\ Head} & \makecell{Keypoint \\ Detection \\ Head} & \makecell{Line \\ Detection \\ Head} & \makecell{Event \\ Classification \\ Head} & \makecell{Vision- \\ Language \\ Alignment} \\
    \midrule
    \textbf{Learning Rate} & $5.0 \times 10^{-5}$ & $1.0 \times 10^{-4}$ & $1.0 \times 10^{-4}$ & $1.0 \times 10^{-4}$ & $1.0 \times 10^{-3}$ & $1.0 \times 10^{-4}$ & $1.0 \times 10^{-4}$ \\
    \bottomrule
    \end{tabular}
\end{table}

\begin{table}[t]
    \centering
    \caption{
        \textbf{Loss Weights for Multi-task Training.}
    }
    \label{tab:loss_weights}
    \footnotesize
    \begin{tabular}{l|c|l}
    \toprule
    \textbf{Loss Component} & \textbf{Weight} & \textbf{Scope} \\
    \midrule
    \rowcolor{gray!20}
    \multicolumn{3}{c}{\textit{Athlete Detection Sub-losses}} \\
    \midrule
    Classification Loss & $\lambda_{\mathrm{cls}} = 2.0$ & Within detection head \\
    Bounding Box Regression Loss & $\lambda_{\mathrm{bbox}} = 5.0$ & Within detection head \\
    Role Classification Loss & $\lambda_{\mathrm{r}} = 2.0$ & Within detection head \\
    Jersey Number Recognition Loss & $\lambda_{\mathrm{j}} = 1.0$ & Within detection head \\
    \midrule
    \rowcolor{gray!20}
    \multicolumn{3}{c}{\textit{Task-level Losses}} \\
    \midrule
    Athlete Detection & $\lambda_{\mathrm{a}} = 1.0$ & Task-level \\
    Keypoint Detection & $\lambda_{\mathrm{k}} = 1.0$ & Task-level \\
    Line Detection & $\lambda_{\mathrm{l}} = 1.0$ & Task-level \\
    Event Classification & $\lambda_{\mathrm{e}} = 1.0$ & Task-level \\
    Vision-Language Alignment & $\lambda_{\mathrm{con}} = 4.0$ & Task-level \\
    \bottomrule
    \end{tabular}
\end{table}

\begin{table}[t]
\centering
    \caption{
        \textbf{Data Augmentation Configuration.} 
        Here, $\pm$ denotes symmetric ranges~({\em e.g.}, $\pm 10$° represents [-10°, 10°]).
    }
    \label{tab:augmentation}
    \footnotesize
    \begin{tabular}{l|c|l}
    \toprule
    \textbf{Augmentation Type} & \textbf{Probability} & \textbf{Parameters} \\
    \midrule
    \rowcolor{gray!20}
    \multicolumn{3}{c}{\textit{Geometric Transformations}} \\
    \midrule
    Random Affine & 0.5 & Rotation: $\pm 10$°, Translation: $\pm 10\%$, Scale: [0.9, 1.1], Shear: $\pm 5$° \\
    Random Perspective & 0.5 & Distortion scale: 0.3 \\
    Random Crop & 1.0 & Size ratio: [0.6, 1.0] \\
    Horizontal Flip & 0.5 & -- \\
    \midrule
    \rowcolor{gray!20}
    \multicolumn{3}{c}{\textit{Photometric Transformations}} \\
    \midrule
    Color Jitter & 1.0 & \makecell[l]{Brightness: 0.4, Contrast: 0.4 \\ Saturation: 0.4, Hue: 0.2} \\
    Gaussian Noise & 0.3 & Standard deviation: 0.02 \\
    Gaussian Blur & 0.2 & Kernel size: [3, 7], Sigma: [0.1, 2.0] \\
    \bottomrule
    \end{tabular}
\end{table}

As specified in Tab.~\ref{tab:augmentation}, we employ a data augmentation strategy incorporating both geometric and photometric transformations to improve model robustness to varying viewpoints, lighting conditions, and image quality.
All pretraining experiments are conducted on $16\times$ NVIDIA H800 GPUs, utilizing a global batch size of 16 for spatial perception tasks and 32 for semantic reasoning tasks. 
The complete multi-task pretraining requires approximately 9 days in full precision~({\em float32}).

\subsection{Implementation details of SoccerFactory}

The field keypoint and line detectors, along with the StrongSORT~\cite{du2023strongsort} and PRTReID~\cite{prtreid} components, are off-the-shelf models utilized in accordance with the SoccerNet-GSR~\cite{SoccerNet-GSR} baseline; specifically, the former use the {\em SV\_kp} and {\em SV\_lines} weights from~\cite{gutierrez2024pnlcalib}, while the latter adopt weights and hyperparameters fine-tuned on the SoccerNet-tracking dataset~\cite{SoccerNetv3-Tracking} from~\cite{prtreid}. For detection, we fine-tune YOLOv8x6~\cite{varghese2024yolov8} on the SoccerNet-GSR training set. For attribute recognition, we leverage Qwen2.5-VL~\cite{Qwen2.5-VL}: the 7B variant is used for athlete role classification, and the 72B variant is employed for jersey number recognition to ensure superior performance, with post-hoc filtering performed by an off-the-shelf legibility classifier~\cite{JerseyNumber} using a confidence threshold of 0.5.

\section{Details of Camera Calibration Metrics}
\label{sec:camera_calibration_metrics}
In our camera calibration evaluation, we strictly adhere to the evaluation protocol in previous works~\cite{gutierrez2024pnlcalib, SoccerNet2022, CameraCalibration}, and report the Jaccard index~($\mathrm{JaC}_\gamma$) at thresholds of 5, 10, and 20 pixels, completion rate~(CR) and final score~(FS).
Specifically, the Jaccard index~($\mathrm{JaC}_\gamma$) measures calibration accuracy based on reprojection error; it is calculated as TP/(TP+FN+FP), where a pitch line is deemed correctly detected only if all its points exhibit reprojection errors below the threshold $\gamma$.
The completion rate~($\mathrm{CR}$) quantifies the proportion of images for which the method successfully produces camera parameters, while the final score, defined as $\mathrm{FS} = \mathrm{CR} \times \mathrm{JaC}_5$, serves as the primary evaluation criterion.

\begin{table*}[t]
    \caption{
        \textbf{Ablation Study on the Impact of Multi-task Pretraining.}
    }
    \centering
    \renewcommand{\arraystretch}{1.1} 
    \resizebox{\linewidth}{!}{
        \begin{tabular}{c | cccc | ccc | ccc | c | c}
        \toprule
        \multirow{2}{*}{\textbf{Multi-task}} & \multicolumn{4}{c|}{\textbf{Athlete Detection}} & \multicolumn{3}{c|}{\textbf{Keypoints Detection}} & \multicolumn{3}{c|}{\textbf{Lines Detection}} & \textbf{Event Classification} & \textbf{Alignment} \\
        \cline{2-13}
         & AP@50 & mAP & jn & role & accuracy & precision & recall & accuracy & precision & recall & Accuracy & Top-1 \\
        \midrule
          & \textbf{88.3} & \textbf{43.8} & \textbf{78.5} & \textbf{98.7} & 93.0 & 90.0 & \textbf{70.9} & \textbf{93.9} & \textbf{96.2} & \textbf{85.5} & 69.1 & 32.6 \\
        \Checkmark & 82.0 & 37.5 & 76.5 & 98.1 & \textbf{93.1} & \textbf{90.1} & 70.7 & 93.7 & 96.1 & 85.2 & \textbf{70.5} & \textbf{36.8} \\
        \bottomrule
        \end{tabular}
    }
    \label{tab:ablation_multitask}
    \vspace{-6pt}
\end{table*}

\section{Ablation on Multi-task Pretraining}
\label{sec:ablation_on_multi-task_pretraining}
To investigate the impact of multi-task pretraining on model performance across different tasks, we conduct an ablation study by comparing two training strategies: 
(i) single-task training, where each task is learned independently, 
and 
(ii) multi-task training, where all tasks are jointly optimized. 
To ensure computational efficiency, both experiments are conducted using our compact SoccerMaster variant.

According to the results presented in Tab.~\ref{tab:ablation_multitask}, we draw the following observations: 
(i) For keypoint detection, line detection, and event classification, both strategies yield comparable performance~(within a 1\% margin).
(ii) However, multi-task pretraining substantially improves vision-language alignment~(top-1: 32.6\% $\rightarrow$ 36.8\%), indicating that joint optimization provides complementary supervisory signals for semantic understanding. 
(iii) Conversely, athlete detection suffers noticeable degradation~(AP@50: 88.3\% $\rightarrow$ 82.0\%, mAP: 43.8\% $\rightarrow$ 37.5\%), which we attribute to the limited capacity of the compact variant struggling with large-scale multi-task optimization.

To validate this hypothesis, we train our full-scale SoccerMaster exclusively on athlete detection and identification, and compare it with the multi-task variant shown in Tab.~3 in the main paper. 
The single-task model yields performance levels comparable to our multi-task model ($-0.3\%$ AP@50, $-0.5\%$ mAP), while showing marginal improvements in jersey number accuracy ($+0.8\%$) and role accuracy ($+0.1\%$).
These negligible fluctuations confirm that, given sufficient model capacity, multi-task pretraining can achieve comparable performance on individual tasks while simultaneously learning robust representations for diverse downstream applications.

\section{Qualitative Results}
\label{sec:qualitative_results}
\subsection{Qualitative Results of SoccerFactory}
\begin{figure*}[t]
    \centering
    \includegraphics[width=0.87\linewidth]{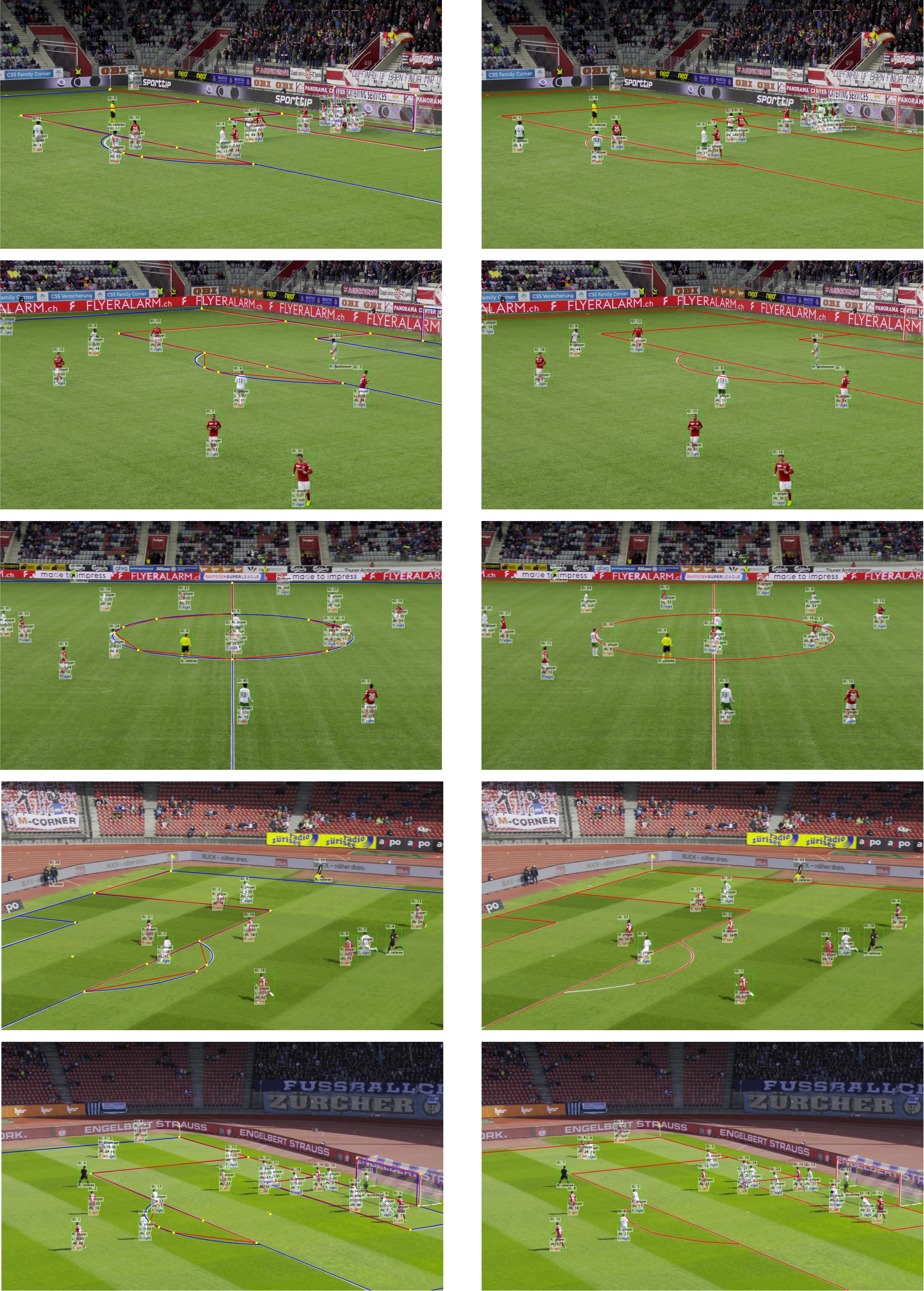}
    \caption{
      \textbf{Qualitative Results of SoccerFactory.} 
      Comparison between our predictions~(left) and ground truth annotations~(right) on the SoccerNet-GSR test set. 
      Our pipeline demonstrates robust performance across diverse scenarios.
    }
    \label{fig:qualitative_pipeline}
\end{figure*}

Fig.~\ref{fig:qualitative_pipeline} presents a visualization example of SoccerFactory on the SoccerNet-GSR~\cite{SoccerNet-GSR} test set, comparing predictions~(left) against ground truth annotations~(right). 
Each detected athlete is annotated with five attributes: 
ID~(tracklet identity), R~(role), L~(legibility score), JN~(jersey number), and T~(team affiliation). 
Regarding field registration, detected keypoints and field lines are highlighted in yellow and red, respectively, while lines projected from the canonical pitch coordinate system via estimated camera parameters are shown in blue. 
In the ground truth images, manually labeled field lines are displayed in red.

These results demonstrate that SoccerFactory achieves robust performance across diverse scenarios. 
The system maintains high accuracy in role classification, jersey number recognition, and team affiliation, even under challenging conditions such as crowded scenes and varying camera viewpoints. 
Notably, SoccerFactory exhibits strong temporal consistency in tracking. 
Specifically, the first and third rows of Fig.~\ref{fig:qualitative_pipeline} correspond to the first and last frames of the same video clip, respectively. 
Despite significant camera motion and player movement across frames, SoccerFactory successfully maintains consistent identity assignments for the tracked athletes, highlighting the effectiveness of our tracking and post-processing modules.

\begin{figure*}[t]
    \centering
    \includegraphics[width=.92\linewidth]{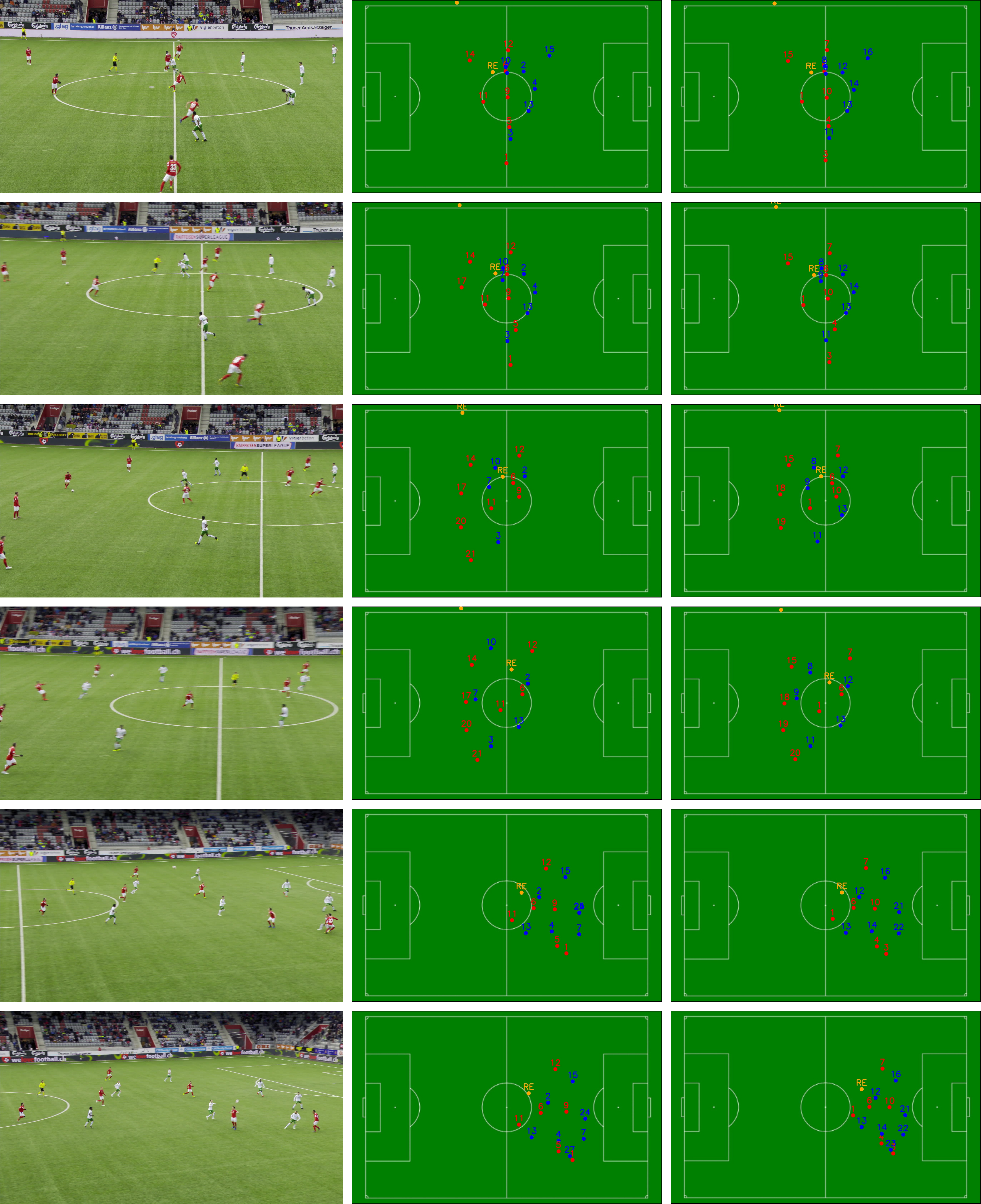}
    \caption{
      \textbf{Top-view Pitch Visualization of SoccerFactory Results.} 
      Athlete positions are mapped to standardized pitch coordinates via estimated camera parameters.
      Each row is organized as: input image~(left), our predictions~(middle), and ground truth annotations~(right).
      Athletes are color-coded by role: referees~(orange, labeled ``RE''), left team~(red), and right team~(blue).
      Non-referee athletes are labeled with arbitrary \textbf{tracklet identities}, which maintain temporal consistency within each tracking sequence.
      Notably, the number of tracklet identities in the predictions is not directly comparable to those in the ground truth; the emphasis is on consistent identity assignment across frames.
    }
    \label{fig:qualitative_pipeline_topview}
\end{figure*}

Fig.~\ref{fig:qualitative_pipeline_topview} presents athlete positions mapped from image coordinates to standardized pitch coordinates using the estimated camera parameters.
The visualization displays a bird's-eye view of the pitch, where the bottom-center of each athlete's bounding box is projected onto the 2D pitch coordinate system to represent their ground-plane position.
Athletes are color-coded by role: referees~(orange, labeled ``RE''), the left team~(red), and the right team~(blue). 
Non-referee athletes are further distinguished by their tracklet identities.
Note that while specific identity values may differ between predictions and ground truth, the primary evaluation criterion is temporal consistency, {\em i.e.}, ensuring the same athlete retains the same identity across frames.
The close alignment between our predictions and ground truth annotations demonstrates the pipeline's capability to accurately estimate camera parameters and maintain consistent athlete tracking across frames.
This precise localization in pitch coordinates can potentially facilitate downstream applications such as tactical analysis~\cite{wang2024tacticai, zhao2025taceleven}, showcasing the practical value of our automatic data curation pipeline beyond basic detection and tracking tasks.

\subsection{Qualitative Results of SoccerMaster}
We present qualitative results of SoccerMaster on a video clip in Fig.~\ref{fig:qualitative_soccer_master}, demonstrating its comprehensive understanding capabilities across multiple soccer understanding tasks. 
The results for multiple object tracking and commentary generation are obtained with task-specific heads and downstream task adaptation. 
The prediction conventions follow the same format as described in Fig.~\ref{fig:qualitative_pipeline}, with generated commentary displayed at the bottom of each frame.

The model exhibits strong performance across athlete detection, pitch registration, multiple object tracking, event classification, and commentary generation. 
Notably, even in crowded scenes, the system maintains robust detection and tracking capabilities, successfully identifying individual athletes and preserving temporal identity consistency across frames. 
Furthermore, despite minor detection errors in keypoint and line predictions in some frames, the camera calibration remains accurate, as evidenced by the close alignment between the projected pitch lines~(blue) and the actual pitch lines and goal frames visible in the RGB images.
This demonstrates the model's robustness to local prediction errors through the geometric refinement process. 
Overall, these results validate SoccerMaster's capability as a unified foundation model to handle diverse soccer understanding tasks within a single framework.

\section{Limitations \& Future Works}
\subsection{Limitations}
While SoccerMaster demonstrates strong performance across diverse soccer understanding tasks, several limitations remain to be addressed in future work.

\begin{figure*}[t]
    \centering
    \includegraphics[width=.94\linewidth]{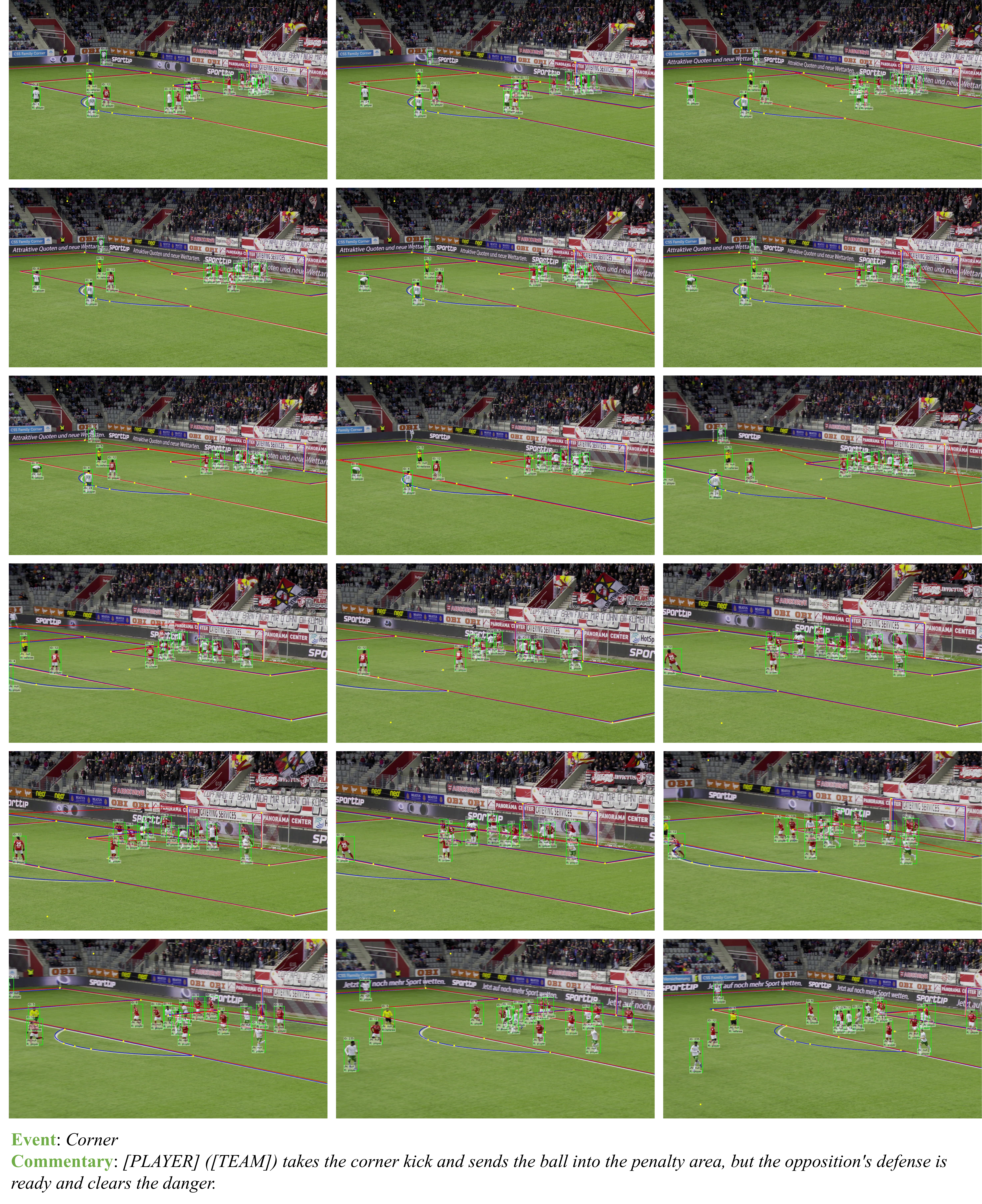}
    \caption{
      \textbf{Qualitative Results of SoccerMaster.} 
      SoccerMaster can simultaneously execute multiple soccer understanding tasks on a video clip, including athlete detection, pitch registration, multiple object tracking, event classification, and commentary generation. 
      Frames are arranged in temporal order from left to right and top to bottom.
    }
    \label{fig:qualitative_soccer_master}
\end{figure*}

\vspace{2pt} 
\noindent \textbf{Jersey number recognition.} 
As illustrated in Fig.~\ref{fig:qualitative_soccer_master}, our model occasionally produces incorrect jersey number predictions. 
We formulate jersey number recognition as a simple 101-class classification problem~(digits 0-99 and \texttt{null}), where the \texttt{null} class dominates the training data due to frequent occlusions and non-frontal viewpoints in broadcast footage. 
This severe class imbalance, combined with limited sample diversity for visible jersey numbers, hinders accurate jersey number recognition in challenging scenarios. 
Furthermore, unlike traditional methods~\cite{JerseyNumber, balaji2023jersey} that apply dedicated recognition models to cropped bounding boxes, our approach performs recognition jointly with detection in a single forward pass. 
While our unified approach offers greater efficiency, the two-stage approach benefits from processing high-resolution crops focused on jersey details, potentially achieving higher accuracy at the cost of increased computational overhead.

\vspace{2pt} 
\noindent \textbf{Goalkeeper classification.} 
Another notable issue is the misclassifications of goalkeepers as players, which can be observed in several frames of Fig.~\ref{fig:qualitative_soccer_master}. 
This problem stems from two primary factors: 
(i) goalkeepers constitute only a small fraction of athletes in the training data, leading to insufficient samples for learning discriminative features; and (ii) goalkeeper uniforms exhibit high variance in color across matches and leagues. 
Distinguishing them from players based solely on individual appearance is difficult without relational reasoning that compares uniform colors across all detected athletes.

\vspace{2pt} 
\noindent \textbf{Limited scope of spatial perception.} 
Our current framework focuses exclusively on athlete detection and pitch registration, without considering ball detection and tracking. 

\subsection{Future Works}
To address the aforementioned limitations and further enhance SoccerMaster's capabilities, our future work will focus on refining the data curation pipeline.
While our current approach is effective, we aim to generate higher-quality annotations at scale by integrating more sophisticated tracking algorithms and enforcing stricter temporal consistency constraints. 
Additionally, we plan to incorporate auxiliary modalities, such as audio commentary and player statistics, to improve annotation accuracy. 
We anticipate that scaling up the training dataset through this enhanced data curation pipeline will yield substantial performance gains across all tasks.

\end{document}